\documentclass[acmtog,screen,balance=true]{acmart}

\copyrightyear{2024} 
\acmYear{2024} 
\setcopyright{acmlicensed}
\acmConference[SA Conference Papers '24]{SIGGRAPH Asia 2024 Conference Papers}{December 3--6, 2024}{Tokyo, Japan}
\acmBooktitle{SIGGRAPH Asia 2024 Conference Papers (SA Conference Papers '24), December 3--6, 2024, Tokyo, Japan}
\acmDOI{10.1145/3680528.3687566}
\acmISBN{979-8-4007-1131-2/24/12}

\usepackage{bm}
\usepackage{subcaption}
\usepackage[capitalize,nameinlink]{cleveref}
\usepackage[export]{adjustbox}
\usepackage{xfrac}
\usepackage{graphicx}
\usepackage{tikz}
\usepackage{xspace}
\usepackage{wrapfig}
\usepackage{tabularx}

\definecolor{myblue}{HTML}{4C72b0}
\definecolor{myorange}{HTML}{DD8452}
\definecolor{myyellow}{HTML}{CC9900}
\definecolor{mygreen}{HTML}{55A868}
\definecolor{myred}{HTML}{C44E52}
\definecolor{mypurple}{HTML}{8172B3}

\newcommand{\sphere}{\Omega}
\newcommand{\latent}{\mathcal{Z}}

\newcommand{\x}{\mathbf{x}}

\newcommand{\pt}{\mathbf{x}}
\newcommand{\dir}{\bm{\omega}}
\newcommand{\wi}{\dir}
\newcommand{\wo}{\dir_\mathrm{o}}

\newcommand{\Lenv}{L}
\newcommand{\Lo}{L_\mathrm{o}}
\newcommand{\normal}{\mathbf{n}}
\newcommand{\brdf}{f_\mathrm{r}^\perp}
\newcommand{\mc}[1]{\langle{#1}\rangle}

\newcommand{\visibility}{V}

\newcommand{\visratio}{\tau}

\newcommand{\pdf}{p}
\newcommand{\target}{\pdf^*}

\newcommand{\uniform}{\mathcal{U}}

\newcommand{\expectation}{\mathbb{E}}

\newcommand{\given}{\,|\,}
\newcommand{\kl}{D_\mathrm{KL}}
\newcommand{\ent}{\mathrm{H}}
\newcommand{\refweight}{\lambda}

\newcommand{\domain}{\mathcal{X}}

\newcommand{\jacobian}{\mathbf{J}}

\newcommand{\flowmodel}{p}
\newcommand{\warp}{f}
\newcommand{\headwarp}{h}
\newcommand{\tailwarp}{t}
\newcommand{\compwarp}{\warp}
\newcommand{\interim}{\mathcal{Y}}

\newcommand{\headwarpparams}{\bm{\theta}}

\newcommand{\allparams}{\bm{\theta}}

\newcommand{\gridfeature}{\bm{\xi}}

\newcommand{\objective}{\mathcal{L}}
\newcommand{\material}{\bm{\rho}}
\newcommand{\condition}{\mathbf{c}}
\newcommand{\roughness}{r}

\newcommand{\scenefmt}[1]{\textsf{\textsc{#1}}}
\newcommand{\templescene}{\scenefmt{Temple}}
\newcommand{\clothscene}{\scenefmt{Cloth}}
\newcommand{\droidscene}{\scenefmt{Droid}}
\newcommand{\dioramascene}{\scenefmt{Diorama}}

\newcommand{\bedscene}{\scenefmt{Bedroom}}
\newcommand{\dishwarescene}{\scenefmt{Dishware}}

\newcommand{\dd}[1]{\mathrm{d}{#1}}
\newcommand{\clamp}[1]{\langle{#1}\rangle}
\newcommand{\dummydot}{\,\cdot\,}

\newcommand{\const}{\mathrm{const.}}

\newcommand{\tikzdiagram}[2]{
\begin{tikzpicture}[every node/.style={inner sep=0,outer sep=0},font=\small\sffamily]
    \node[anchor=south west] (image) at (0,0) {\includegraphics[width=\linewidth]{#1}};
    \begin{scope}[x={(image.south east)},y={(image.north west)}]
        {#2}
    \end{scope}
\end{tikzpicture}}
\begin{document}

\title{Neural Product Importance Sampling via Warp Composition}

\author{Joey Litalien}
    \authornote{Work partly done during an internship at Adobe Research, UK.}
    \email{joey.litalien@gmail.com}
    \orcid{0000-0001-8133-8879}
    \affiliation{\institution{ McGill University} \country{Canada}}
\author{Milo\v{s} Ha\v{s}an}
    \email{mihasan@adobe.com}
    \orcid{0000-0003-3808-6092}
    \affiliation{\institution{Adobe Research} \country{USA}}
\author{Fujun Luan}
    \email{fluan@adobe.com}
    \orcid{0000-0001-5926-6266}
    \affiliation{\institution{Adobe Research} \country{USA}}
\author{Krishna Mullia}
    \email{mulliala@adobe.com}
    \orcid{0000-0002-1892-7299}
    \affiliation{\institution{Adobe Research} \country{USA}}
\author{Iliyan Georgiev}
    \email{igeorgiev@adobe.com}
    \orcid{0000-0002-9655-2138}
    \affiliation{\institution{Adobe Research} \country{UK}}

\renewcommand{\shortauthors}{Litalien, Ha\v{s}an, Luan, Mullia, and Georgiev}

\begin{abstract}
Achieving high efficiency in modern photorealistic rendering hinges on using Monte Carlo sampling distributions that closely approximate the illumination integral estimated for every pixel. Samples are typically generated from a set of simple distributions, each targeting a different factor in the integrand, which are combined via multiple importance sampling. The resulting mixture distribution can be far from the actual product of all factors, leading to sub-optimal variance even for direct-illumination estimation. We present a learning-based method that uses normalizing flows to efficiently importance sample illumination product integrals, e.g., the product of environment lighting and material terms. Our sampler composes a flow head warp with an emitter tail warp. The small conditional head warp is represented by a neural spline flow, while the large unconditional tail is discretized per environment map and its evaluation is instant. If the conditioning is low-dimensional, the head warp can be also discretized to achieve even better performance. We demonstrate variance reduction over prior methods on a range of applications comprising complex geometry, materials and illumination.
\end{abstract}


\begin{teaserfigure}
    \centering
    \includegraphics[width=\linewidth]{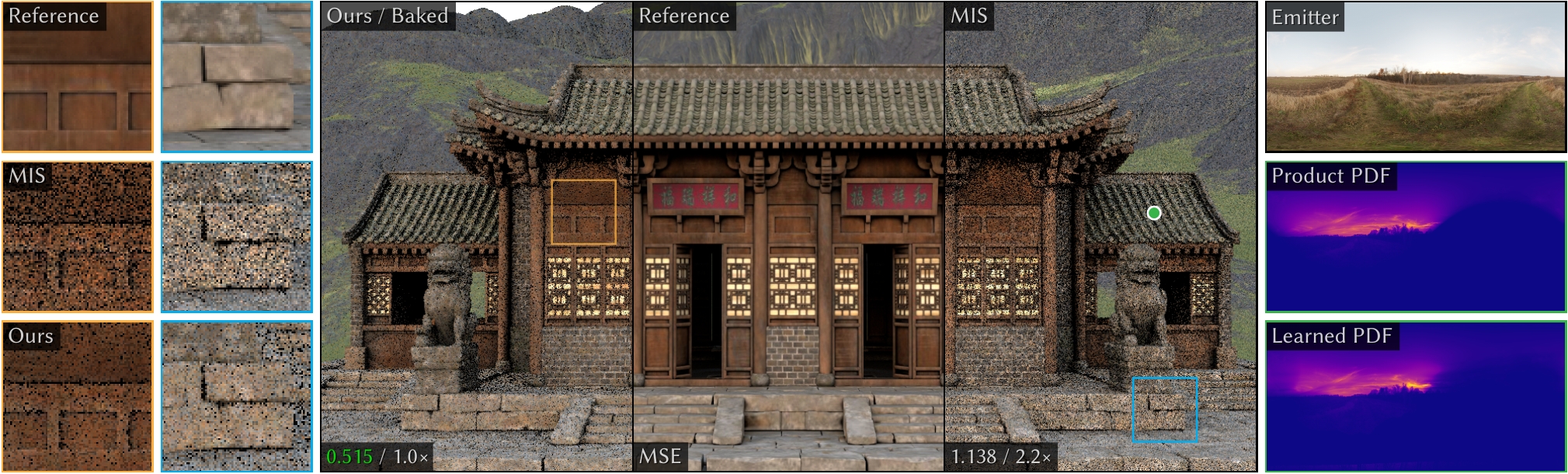}
    \vspace{-6mm}
    \caption{
        Our method composes a neural spline flow \emph{head warp} with an emitter \emph{tail warp} to achieve approximate product importance sampling of environment lighting with other terms (cosine and BRDF). Applied to cosine-weighted environment sampling on the \templescene{} scene, we demonstrate significant variance reduction over multiple importance sampling (MIS) at equal rendering time (35\,ms, 4\,spp). We also visualize the conditional distribution learned by our model at the shading point marked in green. Our learned PDF closely matches the true (unshadowed) product. Our head warp does not have to learn the intricate details of the environment map already captured by the tail warp, and can be represented as a compact normalizing flow that can be baked for fast inference.
    }
    \label{fig:teaser}
    \vspace{3mm}
\end{teaserfigure}

\maketitle

\section{Introduction}

Rendering photorealistic images using Monte Carlo methods requires sampling product integrals with intricate shapes. The most common practical approach is to construct individual estimators that each focus on different factors of the integrand and combine them using multiple importance sampling (MIS) \cite{Veach:1995:MIS}. This combination yields an effective \emph{mixture} sampling distribution; however, it is preferable to approximate the true \emph{product} of the factors, especially in complex material and lighting configurations. Such full product sampling can greatly improve efficiency even the specific case of direct illumination estimation where the product includes the light source, the cosine term, and the material reflectance (BRDF) term.

The rendering community has recently begun investigating importance sampling methods based on neural probabilistic models. Along this axis, discrete normalizing flows (NFs) \cite{Dinh:2014:RealNVP, Rezende:2015:NF} provide an elegant generative framework for constructing flexible distributions by only requiring the specification of a (typically simple) base distribution and a series of bijective transformations, or \emph{warps}. Seminal works \cite{Muller:2019:NIS, Zheng:2019:Learning} showed that NFs can be successfully applied to Monte Carlo rendering by learning on-the-fly from generated samples. However, despite their theoretical appeal and versatility, NF adoption in modern rendering engines has been mostly held back by their high computational cost. Recently, \citet{Xu:2023:NeuSample} have shown that this may not longer hold true, proposing an NF-based framework with practical rendering speed-ups. The ever-growing demand for real-time photorealism has fostered the development of shading languages where neural networks are first-class citizens \cite{Vaidyanathan:2023:Random, Bangaru:2023:SlangD}, which is likely to further improve performance of NFs in future shaders.

We build on these findings and present a method to estimate product integrals using neural probabilistic models. Building atop NFs, our method learns to generate samples from the product distribution of a material model (e.g., a microfacet BRDF) and a distant emitter (an HDR environment map). Key to our method is the composition of a neural spline flow with a fast emitter warp. Our model learns to deform a uniform distribution into an intermediate one that is then transformed to the desired product PDF. We show that imbuing our model with a near-exact emitter warp is an effective inductive bias for neural product sampling. As our head network does not need to learn the fine details of the environment map, it can specialize to focus on the lower-dimensional conditional variations, which drastically simplifies the fitting task.

Our model is compact, shows competitive performance to traditional sampling methods, and integrates easily into an existing rendering pipeline. We implement it into the Mitsuba~3 renderer \cite{Jakob:2022:Mitsuba} and demonstrate reduced variance and improved visual quality on a variety of product sampling applications and scene configurations.

In summary, our key contributions are: 
\begin{itemize}
    \item A new compositional approach for product importance sampling based on normalizing flows, which combines a small but general head warp, represented by a neural spline flow, with a large tail warp, precomputed per environment map for fast evaluation;
    \item a novel neural architecture based on a circular variant of rational-quadratic splines; and
    \item integration into a practical rendering system showing improved performance over previous works in terms of equal sample count and equal time.
\end{itemize}

\section{Background and related Work}

\paragraph{Monte Carlo integration}

We are interested in solving the surface reflection equation \cite{Kajiya:1986:Rendering} which states that the outgoing radiance $\Lo$ at a point $\pt$ in direction $\wo$ is given by
\begin{align}
    \label{eq:directillum}
    \Lo(\pt, \wo) = \int_\sphere \brdf(\pt, \wo, \wi)\,\Lenv(\wi)\,\visibility(\pt,\wi)\,\dd{\wi}.
\end{align}
In this integral over directions $\wi$ on the unit sphere $\sphere$, $\brdf$ is the (cosine-weighted) bidirectional reflectance distribution function (BRDF), $\Lenv$ denotes radiance emitted by a light source which we consider to be distant (i.e., an environment map), and the visibility indicator function $\visibility$ is zero when that radiance is blocked by the scene geometry.

We construct a Monte Carlo (MC) estimator for the above integral:
\begin{align}
    \mc{\Lo}_N = \frac{1}{N} \sum_{i=1}^N 
    \frac{\brdf(\pt, \wo, \dir_i)\,\Lenv(\dir_i)\,\visibility(\pt,\dir_i)}{\pdf(\dir_i\given \x, \wo)},
    \quad
    \dir_i \sim \pdf,
\end{align}
which draws $N$ directions $\dir_i$ from a distribution with probability density function (PDF) $\pdf$. The variance of $\mc{\Lo}_N$ becomes small when $\pdf$ is approximately proportional to the product $\brdf \cdot \Lenv \cdot \visibility$. Finding such a PDF in practice is difficult, and in practice multiple importance sampling (MIS) \cite{Veach:1995:MIS} is used to combine separate estimators of the above form, each using a PDF that targets one of the factors. MIS is suboptimal as it is equivalent to using a mixture sampling distribution (rather than a product) and can be overly defensive \cite{Karlik:2019:MISC}. Devising distributions with closer proportionality to the actual product integrand remains the best way to achieve low estimation variance.

\paragraph{Product importance sampling}

To estimate illumination on glossy surfaces from distant emitters, \citet{Clarberg:2005:Wavelet} proposed a sparse wavelet representation along with a hierarchical sample warping scheme that approximates the product. Given the wavelet's large memory usage, \citet{Clarberg:2008:Practical} instead suggested a sparse quad-tree built on-the-fly from BRDF samples to fit the distribution. \citet{Herholz:2016:Product} estimated indirect illumination via an adaptive Gaussian mixture model optimized via expectation-maximization. \citet{Estevez:2018:Product} used a proxy BRDF representation along with a spherical tabulation technique to perform approximate product importance sampling of multi-lobe materials. These approaches require multiple samples per shading point to amortize the precomputation. In contrast, our model is lightweight and remains efficient in low-sample regimes.

\citet{Xia:2019:Product} proposed a Gaussian representation for product importance sampling of multi-layered materials, while \citet{Villeneuve:2021:VolumeProductSampling} devised analytical techniques for volumetric single scattering along rays. Unlike our framework, these methods are tailored to specific appearances and cannot be easily adapted to account for image-based lighting products.

Resampled importance sampling \cite{Talbot:2005:RIS,Talbot:2005:Thesis,Bitterli:2020:ReSTIR} approximates a target (e.g., product) distribution by carefully picking a subset of candidate samples. The approximation quality is better the closer the candidate distribution is to the target. Resampling can be applied atop our neural sampler.

Closest to our work is the framework of \citet{Hart:2020:Practical} who also recognized that sample warps can---in theory---be composed to achieve perfect product importance sampling. Their work specifically targets area light sources and uses simple analytic warps to approximately correct for the optimal transformation \emph{after} sampling from a known strategy. Our method can be seen as complementary: we \emph{start} with a more flexible, trainable neural warp and then correct with an environment emitter warp. 

\paragraph{Normalizing flows}

Normalizing flows (NFs) are a class of generative models that can construct arbitrary probability distributions \cite{Rezende:2015:NF,Dinh:2014:RealNVP,Chen:2018:NODE}. By design, these models achieve exact likelihood computation and efficient sampling \cite{Papamakarios:2021:NF}, two highly desirable properties for tasks such as density estimation, variational inference and, importantly for us, unbiased integral estimation.

At their core, NFs operate by learning a diffeomorphism (differentiable bijection) $\warp$ from a simple base distribution $\pdf_\latent$ to a more intricate distribution $\pdf_\domain$ via a series of warps that are parameterized by neural networks. Formally, if $\warp\,:\,\latent\,\to\,\domain$ is an invertible transformation with tractable Jacobian determinant $\jacobian_\warp(z;\headwarpparams) \triangleq \det\big(\partial \warp(z;\headwarpparams)/\partial z^T\big)$ and $z \sim \pdf_\latent$, we can express the post-warp density $\pdf_\domain$ via the change-of-variables formula:
\begin{align}
    \label{eq:changeofvar}
    \pdf_\domain(x;\headwarpparams) \, = \, \pdf_\latent(z) \left|\, \jacobian_\warp(z;\headwarpparams) \right|^{-1}, 
\end{align}
where $z = \warp^{-1}(x; \headwarpparams)$, and $\headwarpparams$ represents the parameters of the flow model $\warp$. Under this formulation, efficient sampling of the resulting distribution \emph{and} density evaluation can be simultaneously achieved by mapping $\warp(z; \headwarpparams) = x$ forward and evaluating the determinant along the way. In practice, $\warp$ is implemented as a composition of parameterized \emph{coupling layers}, each designed to have a lower-triangular Jacobian matrix with easy to compute determinant. The parameters $\headwarpparams$ are optimized to approximate a target measure $\target_\domain$ by minimizing a discrepancy metric such as Kullback--Leibler divergence. If samples $x \sim \target_\domain$ are available at training time, this is equivalent to fitting the model by maximum likelihood estimation.

While most NF literature focuses on high-dimensional problems such as image generation \cite{Kingma:2018:Glow, Chen:2019:Residual}, NFs shine in low-dimensional settings where remarkable density fits can be obtained with a fairly low parameter count. To improve robustness, neural spline flows \cite{Durkan:2019:NSF} introduce monotonic rational-quadratic splines to build invertible flow transformations. We adopt the circular variant of this framework \cite{Jimenez:2020:CircularFlows}, with a uniform base distribution, and compose it with other warps to produce high-quality target samples in a stable manner.

\paragraph{Neural sampling for rendering}

Normalizing flows have recently gained prominence in rendering for importance sampling \cite{Muller:2019:NIS,Zheng:2019:Learning}, with extensions to control variates \cite{Muller:2020:NCV}. \citet{Xu:2023:NeuSample} tackled the sampling of neural materials~\cite{Kuznetsov:2021:NeuMIP, Kuznetsov:2022:NeuMIP2} by conditioning an NF model on pretrained neural feature descriptors. \citet{Zeltner:2023:Real} instead utilized a decoder network to extract the parameters of an analytic-lobe mixture for importance sampling complex materials; a similar approach has been explored in prior neural BRDF works~\cite{Ssztrajman:2021:NeuralBRDF, Fan:2022:NeuralLayerBRDF}. Tangentially, \citet {Dong:2023:Mixture} learned neural parametric mixtures for path guiding by implicitly encoding the spatially varying radiance distribution in a scene. Finally, \citet{Vicini:2019:Learned} proposed a BSSRDF model based on a variational autoencoder that learns local geometry with low-order polynomial for subsurface scattering. 

Neural samplers typically learn a full distribution from raw samples and do not leverage the existence of efficient sampling techniques for its individual factors. In contrast, our approach explicitly integrates exact emitter sampling as a form of inductive bias into the pipeline, which drastically simplifies the NF fitting task.

\section{Neural warp composition}
\label{sec:WarpComposition}

Our goal is to generate samples proportionally to the product of two given densities, $\target(x\given\condition) \propto \pdf_1(x) \cdot \pdf_2(x\given\condition)$. We assume that the unconditional density $\pdf_1$ has complex shape (e.g., an unshadowed environment map), while the conditional $\pdf_2$ has simpler shape (e.g., BRDF parameterized by view direction, surface normal, roughness, etc). We do not explicitly model visibility as it is scene-dependent.
 
\begin{figure}[t]%
    \small%
    \setlength{\tabcolsep}{0.4mm}%
    \begin{tabular}{ccccc}%
        \includegraphics[width=0.186\columnwidth,frame]{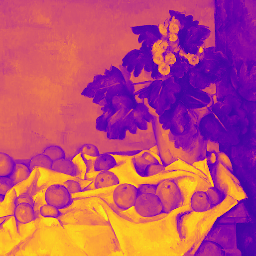}&
        \includegraphics[width=0.186\columnwidth,frame]{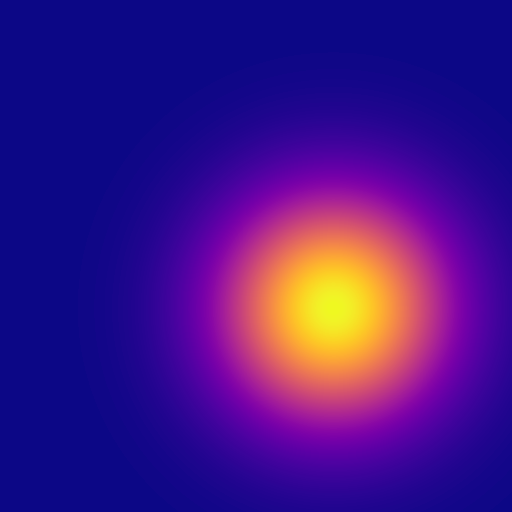}&
        \includegraphics[width=0.186\columnwidth,frame]{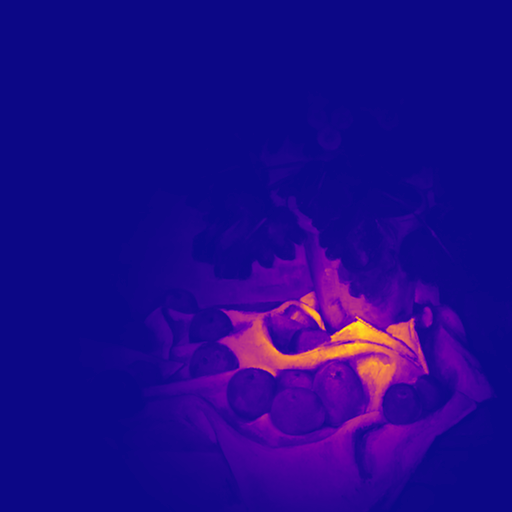}&
        \includegraphics[width=0.186\columnwidth,frame]{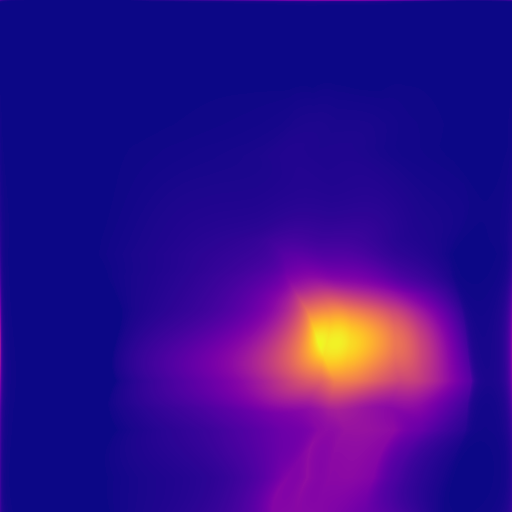}&
        \includegraphics[width=0.186\columnwidth,frame]{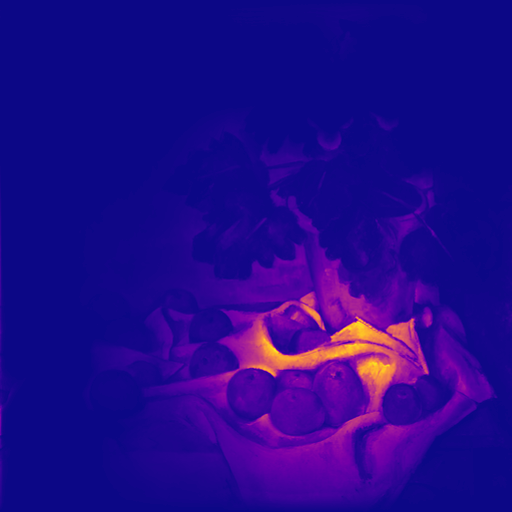}
        \\
        $\pdf_1(\cdot)$ & $\pdf_2(\cdot\given\mu)$ & $\pdf_1 \cdot \pdf_2$ & \textsf{\!\!\!\!Naive product fit\!\!\!\!} & \textsf{\textbf{Our fit}}
    \end{tabular}%
    \vspace{-2mm}
    \caption{
        Naively fitting a normalizing flow (NF) model to the product of a complex unconditioned density $\pdf_1$ (image) and a simple conditioned density $\pdf_2$ (Gaussian with parameterized mean $\mu$) yields a poor result. The model is tasked with simultaneously learning the intricate shape of $\pdf_1$ \emph{and} the variations in $\mu$. Instead, we apply a $\pdf_1$ warp to the NF-model output, which drastically simplifies the shape of the distribution it needs to learn. The result is a near-perfect fit with an equal number of NF parameters.
    }
    \label{fig:2d-example}
    \vspace{-1mm}
\end{figure}

We could train one monolithic flow-based model to warp a base distribution into the target product $\target$. Unfortunately, achieving high run-time performance requires a compact model which would not achieve a good fit because the product has complex shape that varies with the condition $\condition$. We show an example in \cref{fig:2d-example} on a product of an image-based density and a Gaussian with parameterized mean. With this naive strategy, a compact NF model has the capacity to learn only the rough shape of the target distribution.

\begin{figure}[t]
    \centering    
    \tikzdiagram{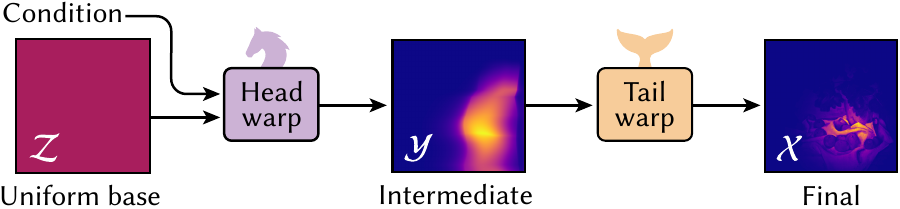}{
        \node[opacity=0.66] at (0.298,0.2) {\cref{sec:head-warp}};
        \node[opacity=0.66] at (0.715,0.2) {\cref{sec:tail-warp}};
    }
    \vspace{-3mm}
    \caption{
        Given a shading condition, our model maps uniform points through two warps to produce samples distributed approximately proportionally to a target product density. The shape of our intermediate density is coarse, similarly to a naive product fit (see \cref{fig:2d-example}), but leads to a precise fit when mapped through the tail warp.
    }
    \label{fig:arch-simplified}
\end{figure}

Our key insight to addressing this problem is that we can handle the complexity and conditioning of the target separately via warp composition. Specifically, instead of using one complex conditional NF model to directly generate samples from the target product $\target$, we use a \emph{compact conditional head} NF model and feed its output to a \emph{complex unconditional tail} warp derived from $\pdf_1$. This separation drastically simplifies the learning space of the NF model by tasking it to fit a smoother intermediate distribution which is subsequently transformed into the final complex product by the tail warp. The latter can be constructed to efficiently handle the intricate shape of $\pdf_1$ as we will discuss below. Note that our head-warp optimization is different from simply fitting to $\pdf_2(\pt\given\condition)$ which would not yield the correct product distribution. \Cref{fig:arch-simplified} shows a high-level overview of our method.

Formally, our model $\compwarp$ is the composition of a conditional NF head warp $\headwarp : \latent\to\interim$, parameterized by $\allparams$, and an unconditional non-parameterized tail warp $\tailwarp: \interim \to \domain$:
\begin{align}
    \compwarp(z \given \condition; \headwarpparams) \triangleq (\tailwarp \circ \headwarp)(z \given \condition; \headwarpparams) = 
    \tailwarp\big(\headwarp(z \given \condition; \headwarpparams)\big).
\end{align}
Provided that $\headwarp$, $\tailwarp$, and their inverses are all continuous, $\compwarp : \latent\to\domain$ is also a diffeomorphism and can be used to fit our target distribution. Hereinafter, we will use subscripts in density notations for disambiguation; our target product density is thus $\target_\domain(x\given\condition)$ and the density learned by our model is $\flowmodel_\domain(x \given \condition;\allparams)$.

\paragraph{Application to rendering}

In our application, the target density is proportional to the product of unshadowed environment light ($\pdf_1$) and cosine-weighted BRDF ($\pdf_2$). The \emph{shading condition} $\condition \triangleq (\wo, \normal, \material)$ includes view direction $\wo$, surface normal $\normal$, and a \linebreak material-specific descriptor $\material$ (e.g., roughness or neural-material embedding). All domains coincide and are square: $\latent = \interim = \domain = [0,1]^2$. The tail-warp output samples $x$ are finally transformed to the unit sphere via a lat-long warp, to obtain a sampled direction $\dir$; we omit that warp here but plot distributions on the domain $\domain$ as lat-long maps.

Next we detail the individual warps and describe how we train their composition via maximum likelihood estimation.

\subsection{Neural-flow head warp} 
\label{sec:head-warp}

The goal of our head warp $\headwarp$ is to transform a base distribution $\flowmodel_\latent$ to an intermediate distribution $\flowmodel_\interim$ that, when pushed through the tail warp $\tailwarp$, yields the target distribution, i.e., $\flowmodel_\domain \approx \target_\domain$. To that end, we build $\headwarp$ upon neural spline flows \cite{Durkan:2019:NSF} which utilize monotonic piecewise rational-quadratic splines as quantile functions (i.e., inverse CDFs) to warp samples. We use the circular variant of these flows \cite{Jimenez:2020:CircularFlows} to produce samples on the unit cylinder, since the azimuth angle in our final lat-long map wraps around. One important distinction over standard NF models is that we use a \textit{uniform} base distribution on the unit square, $\flowmodel_\latent = \uniform[0,1]^2$, to enforce compact support. This is not directly feasible with a normal base distribution.

\begin{figure}[t]
    \centering
    \vspace{-0.6mm}
    \includegraphics[width=\linewidth]{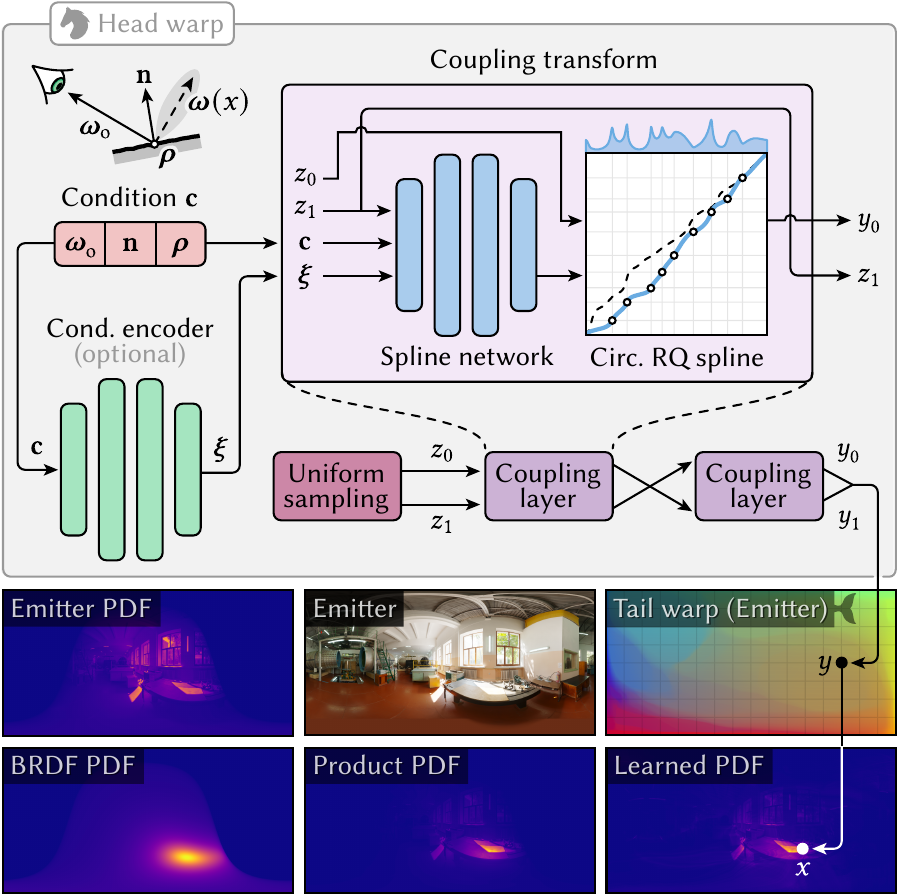}
    \vspace{-5mm}
    \caption{
        Given a {\color{myred}shading condition $\condition$} (view direction $\wo$, surface normal $\normal$ and material descriptor $\material$), a {\color{mygreen}{conditioner encoder}} first produces a latent vector $\gridfeature$. The vectors $\condition$ and $\gridfeature$ condition two {\color{mypurple}{coupling layers}}, each warping samples via a circular piecewise rational quadratic (RQ) spline whose parameters (i.e., knot positions and derivatives) are inferred by a {\color{myblue}{spline network}}. The output $y = (y_0,y_1)$ is then passed through our tail warp to produce the final sample $x$ which is converted to a direction $\dir$ via lat-long mapping.
    }
    \label{fig:arch}
    \vspace{-2mm}
\end{figure}

\Cref{fig:arch} depicts our flow model. Given the shading condition $\condition$, a fully connected encoder network first predicts a feature vector $\gridfeature$ that couples the condition's components in latent space. The concatenation $(\condition, \gridfeature)$ in turn conditions each of two consecutive coupling layers that transform sample coordinates. Each layer has its own network that infers spline parameters. To sample from the model, we draw $z \sim \uniform[0,1]^2$ and feed it to the first coupling layer. The output of the second layer, $\headwarp(z) = y$, is then passed to the next warping stage.

\paragraph{Relation to NeuSample}

Our model is inspired by NeuSample \cite{Xu:2023:NeuSample} but has some key differences. Apart from the already discussed architectural disparities, such as the warp composition and the uniform base, we use a \emph{global} (cylindrical) equi-rectangular parameterization while NeuSample learns on the \emph{local} shading-point hemisphere (projected onto a square). Our method thus guarantees that every point $y\in\interim$ has a pre-image in $\latent$ by design, which is not the case for NeuSample as it needs to learn the disk boundary and occasionally reject out-of-domain samples.

\subsection{Tail warp}
\label{sec:tail-warp}

The tail of our pipeline is an unconditional transformation of head-warp samples $y \in \interim$ to unit directions $\dir$. The samples $y$ are first mapped to unit-square points $x \in \domain$ according to the density defined by a high-dynamic-range environment image (\cref{fig:arch}, bottom right), which are finally lat-long projected to the sphere. Several options for this tail warp are available, such as the common marginal row-column scheme or the hierarchical one of \citet{Clarberg:2005:Wavelet}. Both can be constructed quickly but are discontinuous. This is usually not a major problem in practice, except that discontinuities may ruin the stratification of the input samples $z$. In our case a discontinuous tail warp may hinder the head-warp optimization, since small variations in $\interim$ may lead to abrupt changes in $\domain$. A smooth emitter warp is thus desirable. But how does one construct such a warp?

\begin{figure}
    \centering
    \includegraphics[width=\linewidth]{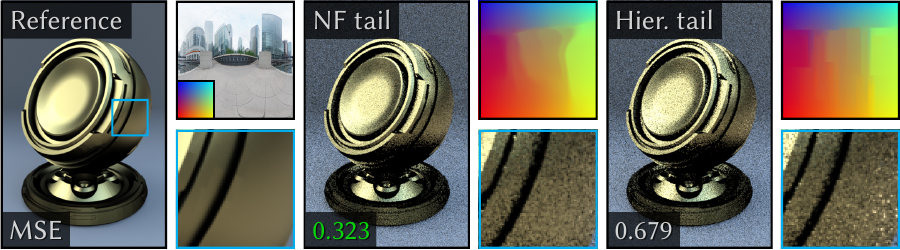}
    \vspace{-6mm}
    \caption{
        Using a hierarchical tail warp exhibits discontinuities and hinders head-warp optimization. We do not show the standard marginal row-column warp as it performs even worse. In contrast, our NF-based tail warp makes for a smooth mapping and halves the MSE of images rendered with our full model. Colors encode the mapping from the unit square.
    }
    \label{fig:emitter-comp}
    \vspace{-3mm}
\end{figure}

We adopt a pragmatic approach and fit a large NF model to the emitter image. By construction, a spline-based flow guarantees a smooth map, and since this specific distribution is unconditional, it can be trained efficiently with samples generated using any of the two aforementioned schemes. Increasing the number of spline bins makes the added approximation error arbitrary small. We considered using an optimal-transport map \cite{Feydy:2019:Sinkhorn,Tong:2024:Improving}. However, due to the regularization required to obtain practical OT solutions, it did not perform well, unless the target itself was smooth (which is far from the case in natural environment maps). \Cref{fig:emitter-comp} shows how the tail-warp smoothness affects rendering quality.

We train the tail NF model and discretize its forward and inverse warps, along with their respective densities, at high resolution. Sampling and PDF evaluation then reduce to simple bilinearly interpolated lookups. Note that this warp is constructed (in isolation) to transform \emph{uniform} density to the emitter-image density. Next, we optimize the head warp to feed \emph{nonuniform} input such that the output density is proportional to the target product.

\subsection{Head-warp training}

We optimize the head-warp parameters $\allparams$ by minimizing the \emph{forward} Kullback--Leibler (KL) loss $\objective_{\mathrm{KL}}(\allparams)$ between the target density $\target_\domain$ and the fit $\flowmodel_\domain$ (below we omit the condition $\condition$ for brevity):
\begin{align}
    \objective_{\mathrm{KL}}(\allparams) &= \kl\big[\target_\domain(x)\,\big\|\,\flowmodel_\domain(x;\allparams)\big]  \\
    &= -\expectation_{x\sim\target_\domain}\big[ \log \flowmodel_\domain(x;\allparams) \big] + \const \label{eq:kl}  \\
    & = -\expectation_{x\sim\target_\domain}\big[ \log \big|\,\jacobian_{\compwarp}(z;\headwarpparams)\big|^{-1}\big]  + \const,
    \label{eq:klbound}
\end{align}
where we have expanded \cref{eq:changeofvar} on the last line, with $z = \compwarp^{-1}(x; \headwarpparams)$, and used the fact that our base distribution $\flowmodel_\latent$ is uniform. This objective amounts to maximizing the log-likelihood w.r.t.\ $\flowmodel_\domain$ using data samples drawn from the target $\target_\domain$. The sampling is done via on-the-fly tabulation and is detailed in \cref{sec:Results}.

Given that our model is the composition $\warp = \tailwarp \circ \headwarp$, the target samples are mapped backward through the inverse composition $f^{-1} = h^{-1} \circ t^{-1}$; the corresponding Jacobians can be readily evaluated along the way. More precisely, we have
\begin{align}
    \log\big|\,\jacobian_{\warp}(z;\headwarpparams)\big|^{-1}
    &= \log\big|\,\jacobian_\tailwarp(\headwarp(z;\headwarpparams))\, \jacobian_\headwarp(z;\headwarpparams)\big|^{-1} \\
    &= \log\big|\,\jacobian_\tailwarp\big(\headwarp(z;\headwarpparams)\big)\big|^{-1} + \log\big|\,\jacobian_\headwarp(z;\headwarpparams)\big|^{-1} \\
    &= \log\big|\,\jacobian_{\tailwarp^{-1}}(x)\big| + \log\big|\,\jacobian_{\headwarp^{-1}}\big(\tailwarp^{-1}(x);\headwarpparams\big)\big|,
    \label{eq:kljacobian}
\end{align}
where in \cref{eq:kljacobian} we have applied the inverse function theorem. 

\paragraph{Entropic regularization}

KL-divergence optimization is susceptible to producing distributions that are overly aggressive in fitting high-density regions, leaving low density regions under-represented to the extent of producing fireflies in MC estimation. Theoretically $\chi^2$ is a better loss that avoids fireflies, however it tends to be too conservative, producing overall higher noise levels, as also observed by \citet{Muller:2019:NIS}.

To improve robustness, we opt for a simpler approach that adds an entropic regularization term to our objective function \eqref{eq:klbound}:
\begin{align}
    \objective(\allparams) &= \objective_{\mathrm{KL}}(\allparams) + \refweight \objective_\ent(\allparams), \qquad \text{where}\\
    \objective_\ent(\allparams) &= \expectation_{x\sim\target_\domain} [\flowmodel_\domain(x;\allparams)\log \flowmodel_\domain(x;\allparams)].
\end{align}
This regularizer penalizes strong mismatched densities and mitigates degenerate cases that may lead to regions with too high variance responsible for fireflies. We set $\refweight = 0.0001$ for all experiments which we found strikes a good balance between outlier suppression and overall variance reduction.

\paragraph{Discussion}

Our training scheme requires samples $x \sim \target_\domain$ from the target. It is tempting to instead optimize the \emph{reverse} KL divergence where the expectation is taken over \emph{model-generated} samples $x \sim \flowmodel_\domain$. That scheme does not necessitate discretizing the target $\target_\domain$ whose evaluation is required only up to a normalization constant \cite{Papamakarios:2021:NF}. Unfortunately, we found that reverse KL optimization is generally too unstable for training with warp compositions. We hypothesize that better initialization schemes for the head flow may alleviate some of the numerical difficulties.

\section{Applications and results}
\label{sec:Results}

We implemented our method in PyTorch \cite{Pytorch}, with fused neural networks in CUDA \cite{Muller:2021:tiny-cuda-nn}, and integrated it into Mitsuba 3's \cite{Jakob:2022:DrJit,Jakob:2022:Mitsuba} wavefront path-tracing renderer. In this section we evaluate its performance on several applications. We begin by describing our experimental setup.

\paragraph{Model hyperparameters}

Our head warp comprises a small conditioner network followed by 2 coupling layers, which we found strikes a good balance between speed and sample quality. We use 32 spline bins. All networks have two layers with 64 neurons each with leaky ReLU activations, for a total of 36k parameters. Our emitter tail warp is a much larger flow network with 128 bins, 16 coupling layers with 256 hidden neurons, and two residual blocks. This network's forward/inverse maps and PDFs are discretized at 1k resolution which we found to be sufficient even for large environment maps.  

\paragraph{Training}

We train using the AdamW optimizer \cite{Loshchilov:2019:AdamW} with a learning rate $\eta = 0.001$ and hyperparameters $\beta=(0.9, 0.999)$. Every training batch contains 256k samples: 1024 samples $x \sim \target_\domain(\dummydot|\,\condition)$ for each of 256 conditions $\condition$. We train for 10k iterations on an NVIDIA A100 GPU; our most costly training with neural materials converges within 30 minutes, while the cosine-weighted and microfacet applications both take 20 minutes to train on average. The one-time pre-training of the tail warp is achieved in under 25 minutes.

For each condition $\condition = (\wo, \normal, \material)$, the outgoing direction $\wo$ and normal $\normal$ are sampled over the sphere. Since $\wo \cdot \normal < 0$ cannot occur for opaque surfaces, we flip the normal accordingly. The material descriptor $\material$ is roughness for the Trowbridge--Reitz model or a NeuMIP feature vector. For Lambertian BRDFs, the condition is simply $\condition = \normal$. To generate samples for the condition, we evaluate the (unshadowed) integrand in \cref{eq:directillum} on a $512 \times 1024$ lat-long grid and tabulate the resulting luminance values into a CDF.

\begin{figure}
    \centering
    \includegraphics[width=\linewidth]{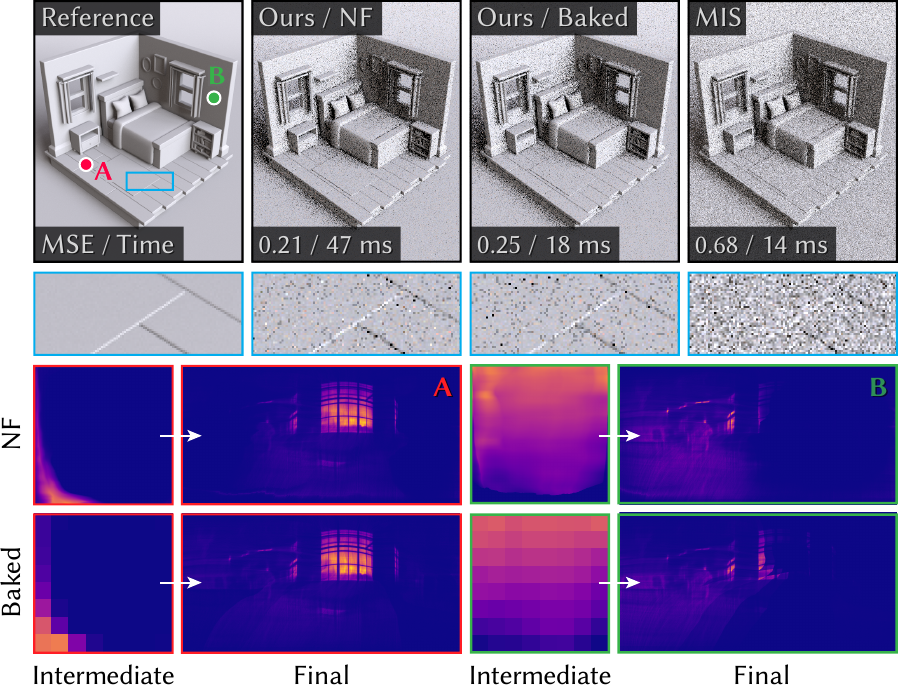}
    \vspace{-4mm}
    \caption{
        Two intermediate distributions (at scene points A \& B) learned by our head-warp NF model (top plots) and their low-resolution bakes (bottom plots) used for fast inference on the \dioramascene scene. Our baked scheme roughly matches the run time of the MIS baseline and achieves a $2.4\times$ MSE reduction over it, with minimal noise increase over the slower NF model. Note that most of the remaining noise for our method comes from visibility.
    }
    \label{fig:histograms-comp}
    \vspace{-1.5mm}
\end{figure}

\paragraph{Evaluation}

We evaluate the performance of our model by measuring the mean square error (MSE) against several baselines at equal rendering time and equal sample count. We found MSE to be less sensitive to outliers than relative MSE in darker regions, which are more omnipresent in direct illumination renderings.

\subsection{Cosine-weighted emitter sampling}

We first demonstrate the benefit of our method to the case of emitter sampling. The popular hierarchical and row-column techniques do not account for surface orientation and generate zero-contribution directions below the local surface horizon, which is suboptimal. Our method can sample the upper hemisphere proportionally to the product of unshadowed radiance and cosine foreshortening by conditioning our model on the surface normal in global space.

\paragraph{Lightweight, baked head warp}

Given the low dimensionality of this problem, we opt for a lighter head-warp architecture without the conditional encoder, directly feeding the normal to the flow network. We also reduce the number of spline bins to 4 and use a hierarchical tail warp as we empirically found that smoothness in the tail warp is not necessary to accurately learn this distribution. This model reduction makes our approach practical, as optimization and tail-warp baking can both be done in under five minutes.

Even with these architectural changes, inference with our neural head warp still incurs a small performance overhead on scenes with simple geometry and materials. The use of MIS with BRDF sampling further exacerbates this problem as our network needs to run twice: once in each direction to compute the MIS weights. To alleviate this issue, we adopt a similar strategy as \citet{Xu:2023:NeuSample} and bake our model into small histograms for fast sampling and PDF evaluation. A key observation here is that we only need to bake the simple \emph{intermediate} distribution (\cref{fig:arch-simplified}) for a set of normal directions. This affords the use of a very low resolution and thus memory footprint. We compute $16 \times 32$ histograms with $8 \times 8$ resolution each (total of 132KB on disk), to achieve a substantial inference speed-up at only slightly increased variance. We label this variant as \emph{Ours / Baked}.
\paragraph{Results}

\begin{figure}
    \centering
    \includegraphics[width=\linewidth]{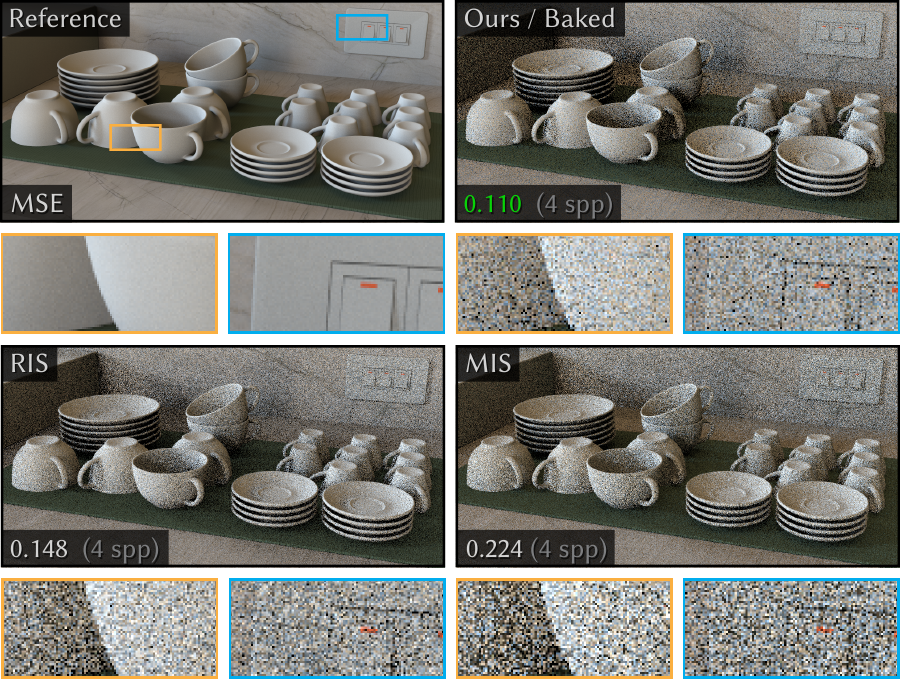}
    \vspace{-5mm}
    \caption{
        \dishwarescene{} scene (equal time). Our compact baked model for cosine-weighted emitter sampling achieves high performance and low variance.
    }
    \label{fig:histograms-comp2}
    \vspace{2mm}
\end{figure}

\Cref{fig:teaser} showcases our cosine-weighted emitter model applied to the \templescene scene. Here, our method can well capture the variations in surface orientations, reducing the variance by $2.2\times$ over MIS across the entire image. Our fast histogram variant achieves equal-time performance at equal sample count. In \cref{fig:histograms-comp}, we visualize intermediate densities of the native NF model and its baked counterpart at two conditions (i.e., surface points). Even at very low resolution, our histograms can yield faithful final product distributions at a fraction of the inference cost. In \cref{fig:histograms-comp2}, we show additional results and compare against MIS and streaming RIS (4 candidates) at equal time, demonstrating once again superior quality over these baselines.

\subsection{Microfacet materials}

As a second application, we fit the product of a distant emitter with a microfacet-based BRDF with a Trowbridge--Reitz (GGX) distribution. The problem is now higher-dimensional as our model has to learn the complex 5D coupling dynamics between surface normal, view direction, and roughness---the conditions to our model.

\begin{figure*}
    \centering
    \includegraphics[width=\linewidth]{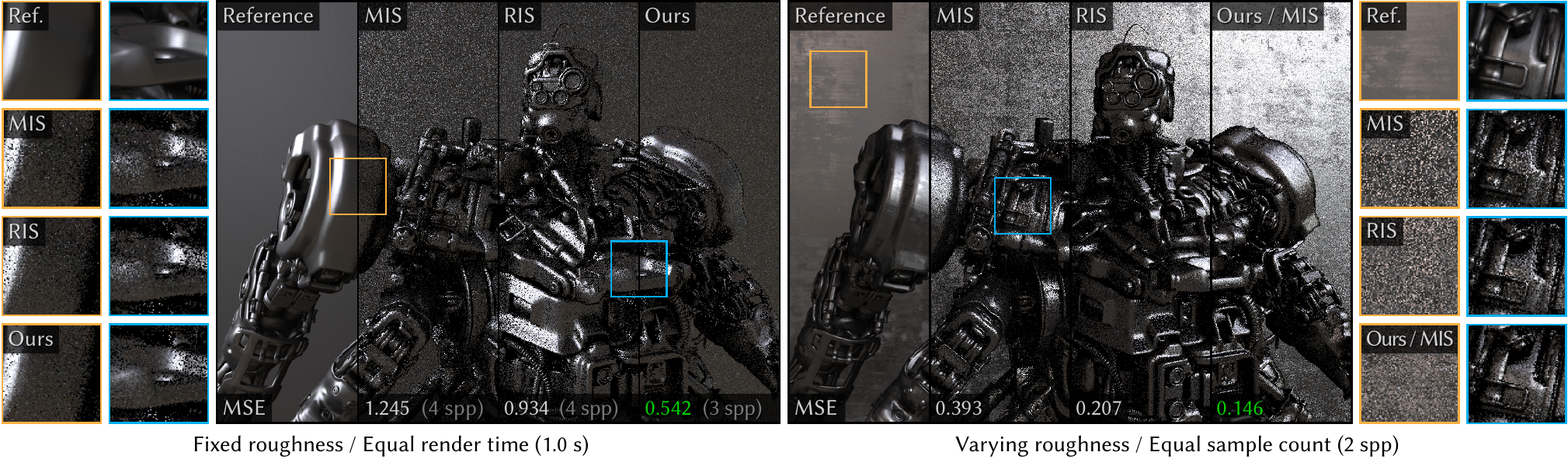}
    \vspace{-6mm}
    \caption{
        \droidscene{} scene. We apply our method to product importance sampling of microfacet BRDF and distant emitter. \emph{Left}: When trained on a fixed roughness value, our model easily outperforms MIS and streaming RIS (4 candidates) at equal rendering time. \emph{Right}: When roughness is an additional condition to our network, our head warp attains a worse fit, but an MIS combination with BRDF sampling achieves variance reduction at equal sample count.
    }
    \label{fig:ggx-results}
\end{figure*}

\paragraph{Results}

To assess how our model adapts to directional and scaling changes in the resulting product lobes, we train two variants: one with fixed roughness $\roughness=0.4$ and another with uniformly sampled roughness $\roughness \in[0.2,0.8]$. We benchmark on a scene with intricate geometry to challenge our model under a wide variety of conditions. To test different roughness values, we apply a texture on the material roughness. We compare against classical MIS and streaming RIS \cite{Bitterli:2020:ReSTIR} with 4 candidates as they are the most practical existing strategies in this scenario.

\Cref{fig:ggx-results} shows the results. At fixed roughness, our model achieves excellent variance reduction at equal time. When roughness is also part of the condition, it still outperforms MIS at equal time but not RIS. We report MSE at equal sample count where we MIS-combine our model with BRDF sampling to further reduce noise. It is worth noting that we keep the capacity of our model to a minimum for performance reasons; as such it may not produce latent vectors $\gridfeature$ that are expressive enough to fully capture the input dependencies. Given the small size of our model, we posit that distilling it into lightweight, fixed-roughness sub-models \cite{Hinton:2014:Distillation} could alleviate this problem; we leave such investigation as future work.

\subsection{Neural materials}

\begin{figure*}
    \centering
    \includegraphics[width=\linewidth]{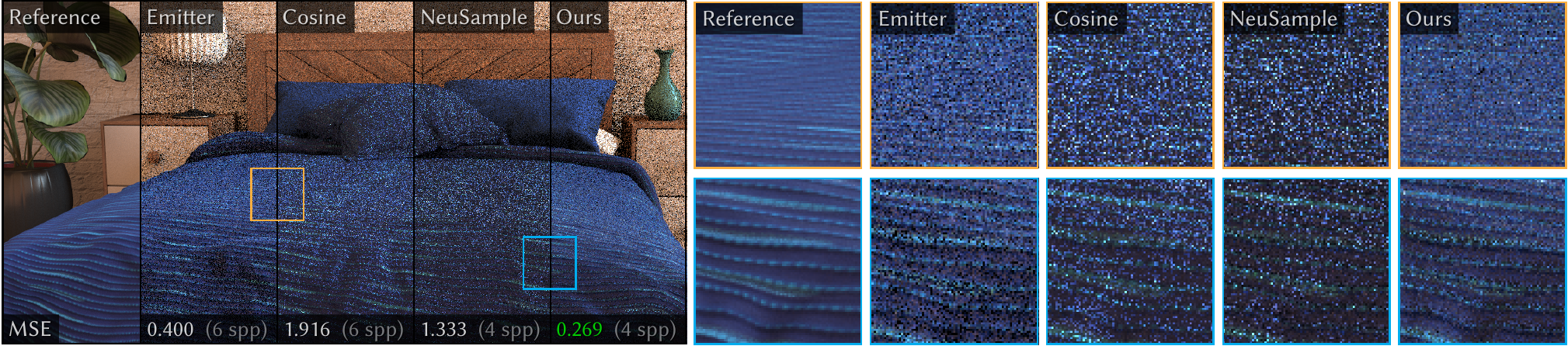}
    \vspace{-6mm}
    \caption{
        \bedscene{} scene (equal time). For high-roughness neural materials like this striped welt fabric, cosine sampling and NeuSample \cite{Xu:2023:NeuSample} both produce high-variance results as they are oblivious to the non-uniform incident radiance. Emitter sampling does better but still samples only one factor in the product. Our model accounts for the product of material and lighting and performs best.
    }
    \label{fig:neural-results1}
\end{figure*}

The application of our sampling framework to products of illumination and neural materials is straightforward. Similarly to NeuSample \cite{Xu:2023:NeuSample}, we use the NeuMIP \cite{Kuznetsov:2021:NeuMIP} feature vector at the shading $uv$-coordinates as the material descriptor $\material$.

\paragraph{Results}

We compare our method against several baseline sampling techniques: cosine-weighted, emitter, and the flow-based variant of NeuSample which we implemented to the best of our ability. NeuSample employs a projected-disk parameterization, whereas our approach is defined in the equirectangular domain and does not suffer from out-of-domain samples---all points in our unit square map to a valid direction. 

\begin{figure*}
\vspace{-1.5mm}
    \centering
    \includegraphics[width=\linewidth]{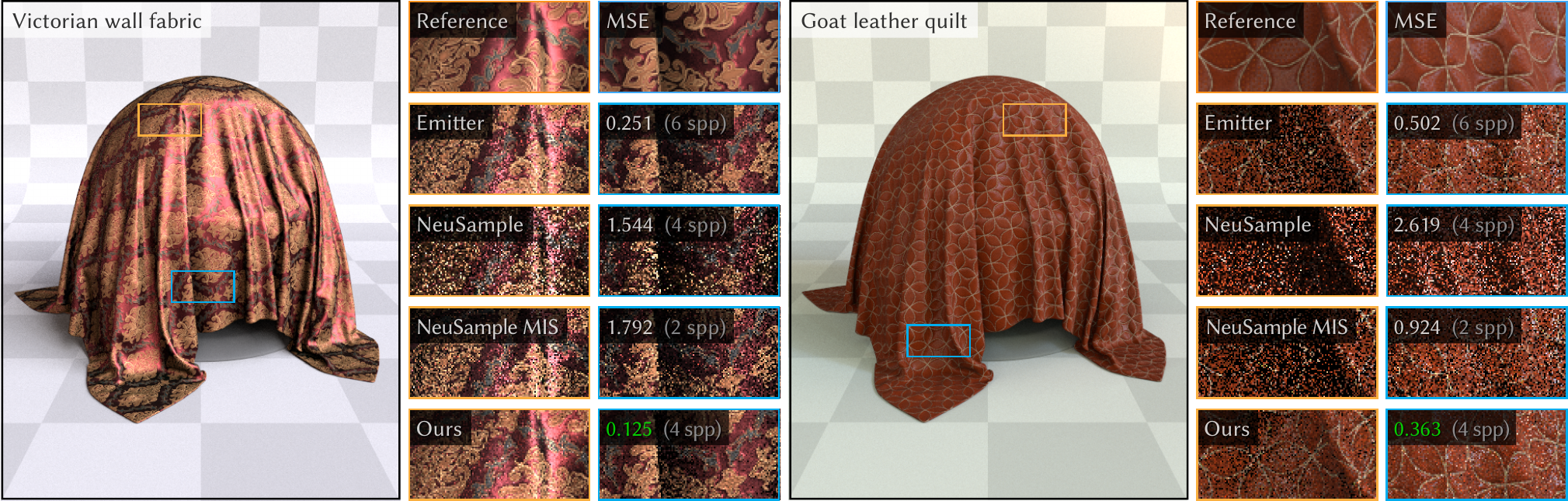}
    \vspace{-6mm}
    \caption{
        \clothscene{} scene (equal time). NeuSample yields excessive noise with lower-roughness neural materials illuminated non-uniformly, even when MIS-combined with emitter sampling. Emitter sampling alone does better but remains inferior to our product approach which generates clear results at equal time.
    }
    \label{fig:neural-results2}
\end{figure*}

We benchmark our approach on three fabric materials with different glossiness. In \cref{fig:neural-results1}, we render a high-roughness striped welt bed sheet and compare against emitter sampling, cosine sampling and NeuSample at equal rendering time. While the neural BRDF exhibits seemingly complex patterns, importance sampling it using NeuSample brings an only small improvement over cosine sampling. Emitter sampling performs favorably to these two but does not account for the full product. Our method performs best as it is informed about both the NeuMIP feature vector and the emitter. \Cref{fig:neural-results2} shows lower-roughness neural materials under directional lighting which prevents NeuSample and its MIS combination with emitter sampling from improving over the baselines; our product-sampling method remains the best by a significant margin.

\subsection{Shadow-catcher compositing}

Finally we show how our model can be employed for compositing virtual objects into photographs using the shadow catcher method. Compositing typically requires two shading passes: one including both the inserted object and the shadow catcher, and one of just the shadow catcher without the object \cite{DebevecDifferential}. We take the ratio of the luminance on the catcher with/without the object and use it as the visibility mask when alpha-composing the object and its shadow into the photograph. This desired transparency (assuming a Lambertian shadow catcher) can be written as a ratio of hemispherical integrals:
\begin{equation}
    \label{eq:ShadowCatcher}
    \visratio = \frac{\int_\sphere \visibility(\x,\wi) \Lenv(\wi) \clamp{\normal, \wi} \,\dd{\wi}}{\int_\sphere \Lenv(\wi) \clamp{\normal, \wi} \,\dd{\wi}},
\end{equation}
where $\Lenv(\wi)$ is the direct luminance (grayscale radiance) from the environment emitter and $\clamp{\normal, \wi}$ is the clamped cosine term. Computing this ratio requires either two rendering passes or significant re-engineering of the renderer to be able to produce these two integral estimates at once.

We note that $\visratio$ is equivalent to the expectation of visibility $\visibility(\x,\wi)$ with respect to a PDF proportional to the product $\Lenv(\wi) \clamp{\normal, \wi}$ (the denominator in \cref{eq:ShadowCatcher} is the PDF's normalization). This density is precisely what our cosine product sampler learns. We can thus estimate the opacity mask in a \emph{single pass} using the simple estimator
\begin{align}
    \mc{\visratio}_N = \frac{1}{N} \sum_{i=1}^N \visibility(\x,\dir_i), \quad \dir_i \sim \flowmodel_\domain(\wi \given\pt; \allparams).
\end{align}
The final composition then amounts to multiplying the visibility mask to the image and adding the masked object.

\paragraph{Results}

\begin{figure}
    \centering
    \includegraphics[width=\linewidth]{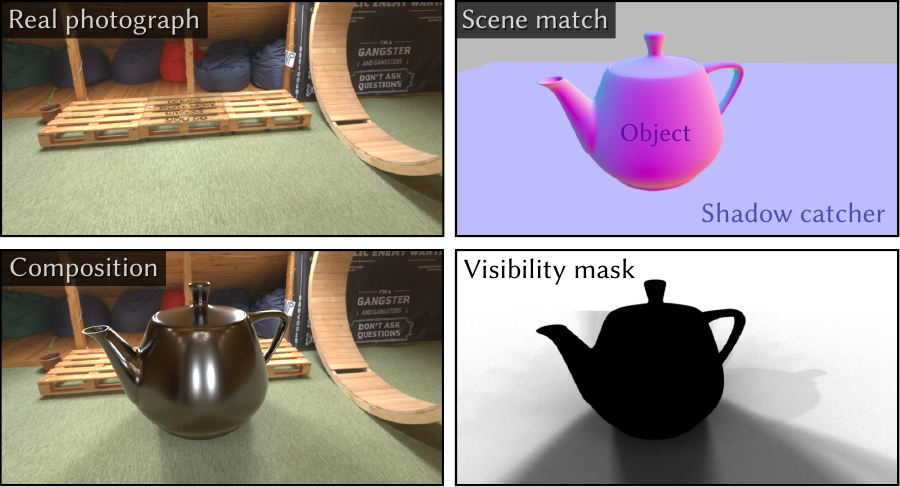}
    \vspace{-6mm}
    \caption{
        Our cosine product sampler can be readily applied to shadow-catcher insertion of virtual objects into real photographs, in one render pass.
    }
    \label{fig:shadow-catcher}
\end{figure}

We demonstrate this application by extracting a rectified image from an environment map to ensure almost exact lighting and use fSpy \cite{Gantelius:2019:fSpy} to roughly match the camera settings and model the shadow receiver plane. We show the resulting composition in \cref{fig:shadow-catcher}. Thanks to our ratio estimator, we get photorealistic object insertion in one rendering pass.

\section{Discussion}

\begin{figure*}
    \centering
    \includegraphics[width=\linewidth]{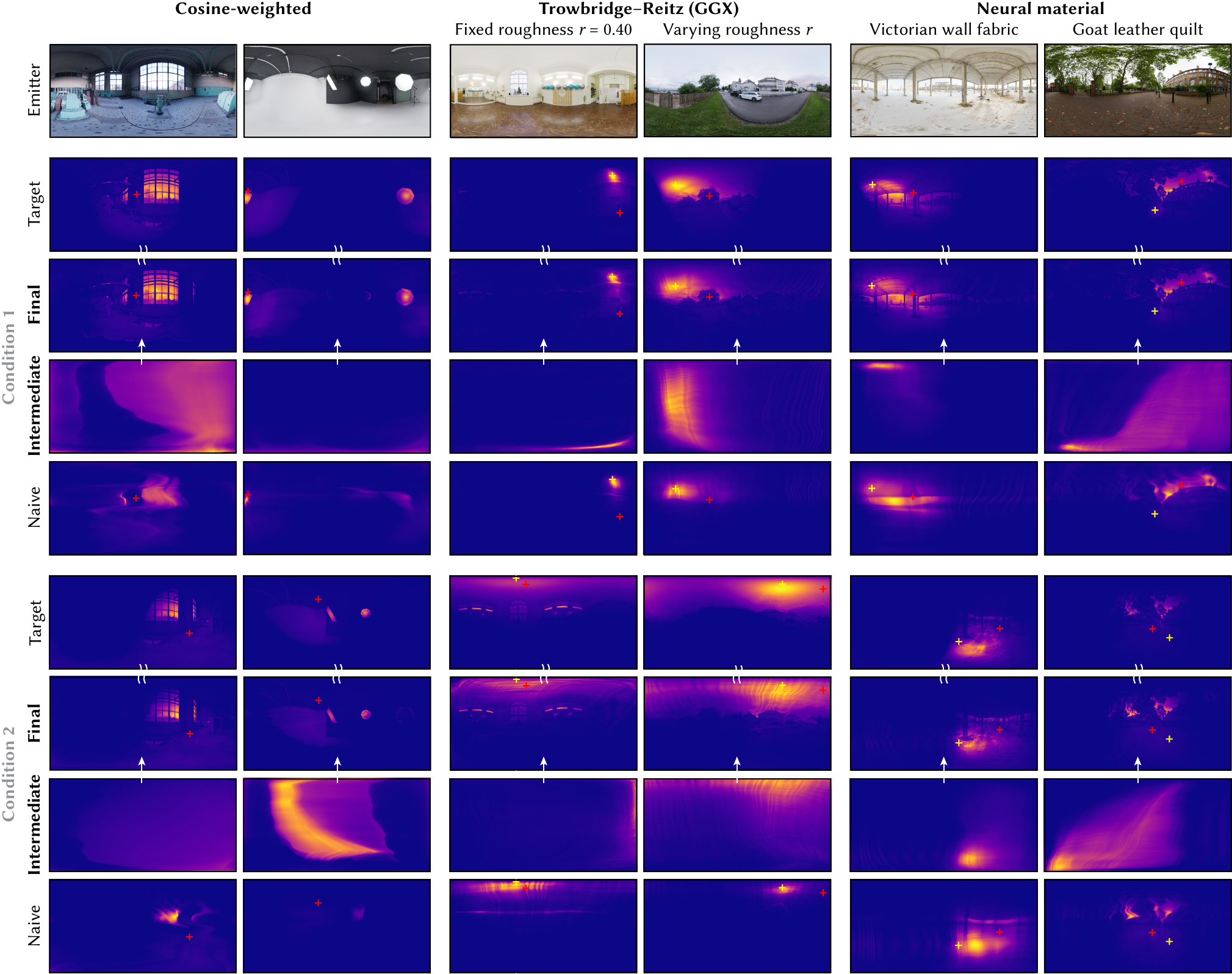}
    \vspace{-6mm}
    \caption{
        For six environment maps, we plot fits of the product with one of three BRDFs used in our experiments. For each of the six configurations, we randomly sample two conditions and show the target product density and the \textbf{intermediate} and \textbf{final} densities learned by our model. For comparison we also visualize a naive neural-flow fit (i.e., without a tail warp). We mark the conditional (global) {\color{myred}{surface normal with a red cross}} and the {\color{myyellow}{reflected view direction with a yellow cross}} to localize the lobe in lat-long coordinates. All our learned final distributions align well with the target product, showcasing the benefit of warp composition for product importance sampling.
    }
    \label{fig:all-fits-results}
    \vspace{-5mm}
\end{figure*}

In \cref{fig:all-fits-results} we plot the intermediate and final densities learned by our model on environment maps and materials used in our experiments. We compare to naively fitting the full product with a neural flow of the same capacity as our head warp. Unlike the naive fits, our final fits align well with their corresponding target densities, showcasing the benefit of separately handling detail and conditioning via warp composition.

\paragraph{Ablation study}

To demonstrate the benefits of the individual components of our model, we conduct an ablation study and report results in \cref{tab:ablation}. Unsurprisingly, naively fitting a flow to a product distribution performs much worse than our two compositional variants. Interestingly, the impact of using the smoother flow-based tail warp over a hierarchical one grows with BRDF complexity. Our conditional encoder and entropic regularization scheme further reduce error. We do not ablate the addition of an encoder for the cosine scheme as it is not part of the model.

\paragraph{Limitations}

Our approach requires training per material model, with duration depending on conditioning dimension and target-distribution complexity. It provides benefit only when the training effort can be amortized in the subsequent rendering. To that end, the total preprocessing time can be cut in half when opting for a standard (e.g., hierarchical) emitter-sampling technique in lieu of optimizing a smooth tail warp. Our cosine-weighted model is particularly practical as it is optimized once per environment map, baked compactly, and used as an efficient drop-in replacement for traditional illumination sampling in any scene (also with MIS).

Our head warp may perform poorly when the product distribution is strongly dominated by the BRDF term (e.g., with very low roughness). We hypothesize that exposing an analytic parameterized BRDF warp to our model may mitigate this issue. Similarly, sun environment maps are not well supported: the tail warp collapses into a single point, which in turn causes numerical issues in the head-warp optimization. MIS can be applied in these scenarios but at the cost of an extra network pass for PDF evaluation. Future hardware acceleration might ease the implementation of fully fused flow kernels to better amortize such additional evaluations.

\begin{table}
    \setlength\tabcolsep{3pt}
    \caption{
        Ablation of the main components of our model on three scenes, reporting rendering MSE. As expected, our full model performs best.
    }
    \label{tab:ablation}
    \vspace{-2.5mm}
    \small
    \begin{tabularx}{\linewidth}{lccc}
        \toprule
        & \textbf{Cosine} & \textbf{Microfacet} & \textbf{\!Neural material\!} \\
        & \templescene{} (\cref{fig:teaser}) & \droidscene{} (\cref{fig:ggx-results}) & \bedscene{} (\cref{fig:neural-results1}) \\ 
        \midrule
        Naive neural-flow fit & 0.715 & 1.411  & 1.320 \\
        Hierarchical tail warp\!\!\!\! & 0.437  & 0.940 & 0.896 \\
        Neural-flow tail warp & 0.406  & 0.901  & 0.739  \\
        $+$ Conditional encoder\!\! & ---  & 0.580  & 0.323 \\
        $+$ Entropic regularization\!\!\!\!\!\!  & \textbf{0.401} & \textbf{0.542}  & \textbf{0.269}  \\
        \bottomrule
    \end{tabularx}
    \vspace{-1mm}
\end{table}

\paragraph{Future work}

Our model is most effective with non-trivial scene complexity, i.e., when sampling is not the principal bottleneck in rendering. Applying our technique to more complex BRDFs, e.g., of layered materials \cite{Guo:2018:Layered}, is thus a promising future avenue. Moreover, our current neural-material approach assumes the existence of a pre-trained (NeuMIP) material model; joint end-to-end training of material and sampling models would be preferable. Making our model MIS-aware could also help mitigate the drawbacks of poor conditional fits for near-unimodal products. For instance, using our model as the free strategy in MIS compensation \cite{Karlik:2019:MISC} could help further reduce variance. Finally, an application to non-axis-aligned portal-masked emitter sampling \cite{Bitterli:2015:Portals} seems within reach, where the flow network could be conditioned on a spatial feature grid to encode portal visibility.

\section{Conclusion}

Efficient product sampling of illumination and reflectance has been a long-standing problem in rendering. We introduced a novel compositional scheme that decouples the handling of distribution shape complexity and conditioning, to afford a compact model capable of achieving both tight fits to target product distributions and fast inference. We demonstrated the versatility of our approach through several practical applications, showing significant variance reduction on scenes of varying complexity. Our method excels when the shape of the product distribution does not degenerate into a singular island of density (e.g., very low roughness BRDFs or sky sun HDRIs). We hope our compositional approach stimulates renewed interest in flow-based models for light-transport simulation, which have been deemed impractical due to their high computational cost.

\begin{acks} 

We  thank the anonymous reviewers for their helpful feedback. All environment HDRIs were obtained from \href{https://polyhaven.com/hdris}{Poly Haven}, while the 3D assets used to model the \templescene{}, \droidscene{}, \dishwarescene{} and \bedscene{} scenes were acquired from \href{https://www.blenderkit.com/}{BlenderKit}. 

\end{acks}

\bibliographystyle{ACM-Reference-Format}
\bibliography{references}


\begin{thebibliography}{46}


\ifx \showCODEN    \undefined \def \showCODEN     #1{\unskip}     \fi
\ifx \showDOI      \undefined \def \showDOI       #1{#1}\fi
\ifx \showISBNx    \undefined \def \showISBNx     #1{\unskip}     \fi
\ifx \showISBNxiii \undefined \def \showISBNxiii  #1{\unskip}     \fi
\ifx \showISSN     \undefined \def \showISSN      #1{\unskip}     \fi
\ifx \showLCCN     \undefined \def \showLCCN      #1{\unskip}     \fi
\ifx \shownote     \undefined \def \shownote      #1{#1}          \fi
\ifx \showarticletitle \undefined \def \showarticletitle #1{#1}   \fi
\ifx \showURL      \undefined \def \showURL       {\relax}        \fi
\providecommand\bibfield[2]{#2}
\providecommand\bibinfo[2]{#2}
\providecommand\natexlab[1]{#1}
\providecommand\showeprint[2][]{arXiv:#2}

\bibitem[Bangaru et~al\mbox{.}(2023)]%
        {Bangaru:2023:SlangD}
\bibfield{author}{\bibinfo{person}{Sai~Praveen Bangaru}, \bibinfo{person}{Lifan
  Wu}, \bibinfo{person}{Tzu-Mao Li}, \bibinfo{person}{Jacob Munkberg},
  \bibinfo{person}{Gilbert Bernstein}, \bibinfo{person}{Jonathan Ragan-Kelley},
  \bibinfo{person}{Fr\'{e}do Durand}, \bibinfo{person}{Aaron Lefohn}, {and}
  \bibinfo{person}{Yong He}.} \bibinfo{year}{2023}\natexlab{}.
\newblock \showarticletitle{SLANG.D: Fast, Modular and Differentiable Shader
  Programming}.
\newblock \bibinfo{journal}{\emph{ACM Transactions on Graphics}}
  \bibinfo{volume}{42}, \bibinfo{number}{6}, Article \bibinfo{articleno}{264}
  (\bibinfo{year}{2023}).
\newblock
\urldef\tempurl%
\url{https://doi.org/10.1145/3618353}
\showDOI{\tempurl}


\bibitem[Bitterli et~al\mbox{.}(2015)]%
        {Bitterli:2015:Portals}
\bibfield{author}{\bibinfo{person}{Benedikt Bitterli}, \bibinfo{person}{Jan
  Novák}, {and} \bibinfo{person}{Wojciech Jarosz}.}
  \bibinfo{year}{2015}\natexlab{}.
\newblock \showarticletitle{Portal-Masked Environment Map Sampling}.
\newblock \bibinfo{journal}{\emph{Computer Graphics Forum (Proceedings of
  EGSR)}} \bibinfo{volume}{34}, \bibinfo{number}{4} (\bibinfo{date}{June}
  \bibinfo{year}{2015}).
\newblock
\urldef\tempurl%
\url{https://doi.org/10/f7mbx7}
\showDOI{\tempurl}


\bibitem[Bitterli et~al\mbox{.}(2020)]%
        {Bitterli:2020:ReSTIR}
\bibfield{author}{\bibinfo{person}{Benedikt Bitterli}, \bibinfo{person}{Chris
  Wyman}, \bibinfo{person}{Matt Pharr}, \bibinfo{person}{Peter Shirley},
  \bibinfo{person}{Aaron Lefohn}, {and} \bibinfo{person}{Wojciech Jarosz}.}
  \bibinfo{year}{2020}\natexlab{}.
\newblock \showarticletitle{Spatiotemporal Reservoir Resampling for Real-time
  Ray Tracing with Dynamic Direct Lighting}.
\newblock \bibinfo{journal}{\emph{ACM Transactions on Graphics}}
  \bibinfo{volume}{39}, \bibinfo{number}{4} (\bibinfo{year}{2020}).
\newblock
\urldef\tempurl%
\url{https://doi.org/10.1145/3386569.3392481}
\showDOI{\tempurl}


\bibitem[Chen et~al\mbox{.}(2019)]%
        {Chen:2019:Residual}
\bibfield{author}{\bibinfo{person}{Ricky T.~Q. Chen}, \bibinfo{person}{Jens
  Behrmann}, \bibinfo{person}{David Duvenaud}, {and}
  \bibinfo{person}{J\"{o}rn-Henrik Jacobsen}.} \bibinfo{year}{2019}\natexlab{}.
\newblock \showarticletitle{Residual Flows for Invertible Generative Modeling}.
  In \bibinfo{booktitle}{\emph{Proceedings of the 33rd International Conference
  on Neural Information Processing Systems}}.
\newblock
\urldef\tempurl%
\url{https://doi.org/10.5555/3454287.3455176}
\showDOI{\tempurl}


\bibitem[Chen et~al\mbox{.}(2018)]%
        {Chen:2018:NODE}
\bibfield{author}{\bibinfo{person}{Ricky T.~Q. Chen}, \bibinfo{person}{Yulia
  Rubanova}, \bibinfo{person}{Jesse Bettencourt}, {and} \bibinfo{person}{David
  Duvenaud}.} \bibinfo{year}{2018}\natexlab{}.
\newblock \showarticletitle{Neural Ordinary Differential Equations}. In
  \bibinfo{booktitle}{\emph{Proceedings of the 32nd International Conference on
  Neural Information Processing Systems}}.
\newblock
\urldef\tempurl%
\url{https://doi.org/10.5555/3327757.3327764}
\showDOI{\tempurl}


\bibitem[Clarberg and Akenine-M\"oller(2008)]%
        {Clarberg:2008:Practical}
\bibfield{author}{\bibinfo{person}{Petrik Clarberg} {and}
  \bibinfo{person}{Tomas Akenine-M\"oller}.} \bibinfo{year}{2008}\natexlab{}.
\newblock \showarticletitle{Practical Product Importance Sampling for Direct
  Illumination}.
\newblock \bibinfo{journal}{\emph{Computer Graphics Forum}}
  \bibinfo{volume}{27}, \bibinfo{number}{2} (\bibinfo{year}{2008}).
\newblock
\urldef\tempurl%
\url{https://doi.org/10.1111/j.1467-8659.2008.01166.x}
\showDOI{\tempurl}


\bibitem[Clarberg et~al\mbox{.}(2005)]%
        {Clarberg:2005:Wavelet}
\bibfield{author}{\bibinfo{person}{Petrik Clarberg}, \bibinfo{person}{Wojciech
  Jarosz}, \bibinfo{person}{Tomas Akenine-M\"oller}, {and}
  \bibinfo{person}{Henrik~Wann Jensen}.} \bibinfo{year}{2005}\natexlab{}.
\newblock \showarticletitle{Wavelet Importance Sampling: Efficiently Evaluating
  Products of Complex Functions}.
\newblock \bibinfo{journal}{\emph{ACM Transactions on Graphics}}
  \bibinfo{volume}{24}, \bibinfo{number}{3} (\bibinfo{date}{Aug.}
  \bibinfo{year}{2005}), \bibinfo{pages}{1166--1175}.
\newblock
\urldef\tempurl%
\url{https://doi.org/10.1145/1073204.1073328}
\showDOI{\tempurl}


\bibitem[Conty~Estevez and Lecocq(2018)]%
        {Estevez:2018:Product}
\bibfield{author}{\bibinfo{person}{Alejandro Conty~Estevez} {and}
  \bibinfo{person}{Pascal Lecocq}.} \bibinfo{year}{2018}\natexlab{}.
\newblock \showarticletitle{Fast Product Importance Sampling of Environment
  Maps}. In \bibinfo{booktitle}{\emph{ACM SIGGRAPH 2018 Talks}}. Article
  \bibinfo{articleno}{69}.
\newblock
\urldef\tempurl%
\url{https://doi.org/10.1145/3214745.3214760}
\showDOI{\tempurl}


\bibitem[Debevec(1998)]%
        {DebevecDifferential}
\bibfield{author}{\bibinfo{person}{Paul Debevec}.}
  \bibinfo{year}{1998}\natexlab{}.
\newblock \showarticletitle{Rendering Synthetic Objects Into Real Scenes:
  Bridging Traditional and Image-based Graphics with Global Illumination and
  High Dynamic Range Photography}. In \bibinfo{booktitle}{\emph{Proceedings of
  the 25th Annual Conference on Computer Graphics and Interactive Techniques}}
  \emph{(\bibinfo{series}{SIGGRAPH '98})}. \bibinfo{pages}{189–198}.
\newblock
\urldef\tempurl%
\url{https://doi.org/10.1145/280814.280864}
\showDOI{\tempurl}


\bibitem[Dinh et~al\mbox{.}(2014)]%
        {Dinh:2014:RealNVP}
\bibfield{author}{\bibinfo{person}{Laurent Dinh}, \bibinfo{person}{Jascha
  Sohl{-}Dickstein}, {and} \bibinfo{person}{Samy Bengio}.}
  \bibinfo{year}{2014}\natexlab{}.
\newblock \showarticletitle{Density Estimation Using Real {NVP}}. In
  \bibinfo{booktitle}{\emph{International Conference on Learning
  Representations}}.
\newblock


\bibitem[Dong et~al\mbox{.}(2023)]%
        {Dong:2023:Mixture}
\bibfield{author}{\bibinfo{person}{Honghao Dong}, \bibinfo{person}{Guoping
  Wang}, {and} \bibinfo{person}{Sheng Li}.} \bibinfo{year}{2023}\natexlab{}.
\newblock \showarticletitle{Neural Parametric Mixtures for Path Guiding}. In
  \bibinfo{booktitle}{\emph{SIGGRAPH '23 Conference Proceedings}}.
\newblock
\urldef\tempurl%
\url{https://doi.org/10.1145/3588432.3591533}
\showDOI{\tempurl}


\bibitem[Durkan et~al\mbox{.}(2019)]%
        {Durkan:2019:NSF}
\bibfield{author}{\bibinfo{person}{Conor Durkan}, \bibinfo{person}{Artur
  Bekasov}, \bibinfo{person}{Iain Murray}, {and} \bibinfo{person}{George
  Papamakarios}.} \bibinfo{year}{2019}\natexlab{}.
\newblock \showarticletitle{Neural Spline Flows}. In
  \bibinfo{booktitle}{\emph{Proceedings of the 33rd International Conference on
  Neural Information Processing Systems}}. Article \bibinfo{articleno}{675}.
\newblock
\urldef\tempurl%
\url{https://doi.org/10.5555/3454287.3454962}
\showDOI{\tempurl}


\bibitem[Fan et~al\mbox{.}(2022)]%
        {Fan:2022:NeuralLayerBRDF}
\bibfield{author}{\bibinfo{person}{Jiahui Fan}, \bibinfo{person}{Beibei Wang},
  \bibinfo{person}{Milo\v{s} Ha\v{s}an}, \bibinfo{person}{Jian Yang}, {and}
  \bibinfo{person}{Ling-Qi Yan}.} \bibinfo{year}{2022}\natexlab{}.
\newblock \showarticletitle{Neural Layered BRDFs}. In
  \bibinfo{booktitle}{\emph{ACM SIGGRAPH 2022 Conference Proceedings}}.
\newblock
\urldef\tempurl%
\url{https://doi.org/10.1145/3528233.3530732}
\showDOI{\tempurl}


\bibitem[Feydy et~al\mbox{.}(2019)]%
        {Feydy:2019:Sinkhorn}
\bibfield{author}{\bibinfo{person}{Jean Feydy}, \bibinfo{person}{Thibault
  S{\'e}journ{\'e}}, \bibinfo{person}{Fran{\c{c}}ois-Xavier Vialard},
  \bibinfo{person}{Shun-ichi Amari}, \bibinfo{person}{Alain Trouve}, {and}
  \bibinfo{person}{Gabriel Peyr{\'e}}.} \bibinfo{year}{2019}\natexlab{}.
\newblock \showarticletitle{Interpolating between Optimal Transport and {MMD}
  using {Sinkhorn} Divergences}. In \bibinfo{booktitle}{\emph{The 22nd
  International Conference on Artificial Intelligence and Statistics}}.
  \bibinfo{pages}{2681--2690}.
\newblock


\bibitem[Gantelius(2019)]%
        {Gantelius:2019:fSpy}
\bibfield{author}{\bibinfo{person}{Per Gantelius}.}
  \bibinfo{year}{2019}\natexlab{}.
\newblock \bibinfo{booktitle}{\emph{{fSpy}}}.
\newblock
\urldef\tempurl%
\url{https://github.com/stuffmatic/fSpy-Blender}
\showURL{%
\tempurl}


\bibitem[Guo et~al\mbox{.}(2018)]%
        {Guo:2018:Layered}
\bibfield{author}{\bibinfo{person}{Yu Guo}, \bibinfo{person}{Milo\v{s}
  Ha\v{s}an}, {and} \bibinfo{person}{Shaung Zhao}.}
  \bibinfo{year}{2018}\natexlab{}.
\newblock \showarticletitle{Position-Free Monte Carlo Simulation for Arbitrary
  Layered BSDFs}.
\newblock \bibinfo{journal}{\emph{ACM Transactions on Graphics}}
  \bibinfo{volume}{37}, \bibinfo{number}{6} (\bibinfo{year}{2018}).
\newblock
\urldef\tempurl%
\url{https://doi.org/10.1145/3272127.3275053}
\showDOI{\tempurl}


\bibitem[Hart et~al\mbox{.}(2020)]%
        {Hart:2020:Practical}
\bibfield{author}{\bibinfo{person}{David Hart}, \bibinfo{person}{Matt Pharr},
  \bibinfo{person}{Thomas Müller}, \bibinfo{person}{Ward Lopes},
  \bibinfo{person}{Morgan McGuire}, {and} \bibinfo{person}{Peter Shirley}.}
  \bibinfo{year}{2020}\natexlab{}.
\newblock \showarticletitle{{Practical Product Sampling by Fitting and
  Composing Warps}}.
\newblock \bibinfo{journal}{\emph{Computer Graphics Forum}}
  (\bibinfo{year}{2020}).
\newblock
\showISSN{1467-8659}
\urldef\tempurl%
\url{https://doi.org/10.1111/cgf.14060}
\showDOI{\tempurl}


\bibitem[Herholz et~al\mbox{.}(2016)]%
        {Herholz:2016:Product}
\bibfield{author}{\bibinfo{person}{Sebastian Herholz}, \bibinfo{person}{Oskar
  Elek}, \bibinfo{person}{Jiří Vorba}, \bibinfo{person}{Hendrik Lensch},
  {and} \bibinfo{person}{Jaroslav K{\v{r}}iv{\'{a}}nek}.}
  \bibinfo{year}{2016}\natexlab{}.
\newblock \showarticletitle{Product Importance Sampling for Light Transport
  Path Guiding}.
\newblock \bibinfo{journal}{\emph{Computer Graphics Forum}}
  \bibinfo{volume}{35}, \bibinfo{number}{4} (\bibinfo{year}{2016}),
  \bibinfo{pages}{67--77}.
\newblock
\urldef\tempurl%
\url{https://doi.org/10.1111/cgf.12950}
\showDOI{\tempurl}


\bibitem[Hinton et~al\mbox{.}(2015)]%
        {Hinton:2014:Distillation}
\bibfield{author}{\bibinfo{person}{Geoffrey Hinton}, \bibinfo{person}{Oriol
  Vinyals}, {and} \bibinfo{person}{Jeffrey Dean}.}
  \bibinfo{year}{2015}\natexlab{}.
\newblock \showarticletitle{Distilling the Knowledge in a Neural Network}. In
  \bibinfo{booktitle}{\emph{NIPS Deep Learning and Representation Learning
  Workshop}}.
\newblock


\bibitem[Jakob et~al\mbox{.}(2022b)]%
        {Jakob:2022:Mitsuba}
\bibfield{author}{\bibinfo{person}{Wenzel Jakob}, \bibinfo{person}{Sébastien
  Speierer}, \bibinfo{person}{Nicolas Roussel}, \bibinfo{person}{Merlin
  Nimier-David}, \bibinfo{person}{Delio Vicini}, \bibinfo{person}{Tizian
  Zeltner}, \bibinfo{person}{Baptiste Nicolet}, \bibinfo{person}{Miguel
  Crespo}, \bibinfo{person}{Vincent Leroy}, {and} \bibinfo{person}{Ziyi
  Zhang}.} \bibinfo{year}{2022}\natexlab{b}.
\newblock \bibinfo{booktitle}{\emph{Mitsuba 3 Renderer}}.
\newblock
\urldef\tempurl%
\url{https://mitsuba-renderer.org}
\showURL{%
\tempurl}


\bibitem[Jakob et~al\mbox{.}(2022a)]%
        {Jakob:2022:DrJit}
\bibfield{author}{\bibinfo{person}{Wenzel Jakob}, \bibinfo{person}{Sébastien
  Speierer}, \bibinfo{person}{Nicolas Roussel}, {and} \bibinfo{person}{Delio
  Vicini}.} \bibinfo{year}{2022}\natexlab{a}.
\newblock \showarticletitle{Dr.Jit: A Just-In-Time Compiler for Differentiable
  Rendering}.
\newblock \bibinfo{journal}{\emph{ACM Transactions on Graphics}}
  \bibinfo{volume}{41}, \bibinfo{number}{4} (\bibinfo{date}{July}
  \bibinfo{year}{2022}).
\newblock
\urldef\tempurl%
\url{https://doi.org/10.1145/3528223.3530099}
\showDOI{\tempurl}


\bibitem[Kajiya(1986)]%
        {Kajiya:1986:Rendering}
\bibfield{author}{\bibinfo{person}{James~T. Kajiya}.}
  \bibinfo{year}{1986}\natexlab{}.
\newblock \showarticletitle{The Rendering Equation}
  \emph{(\bibinfo{series}{SIGGRAPH '86})}. \bibinfo{pages}{143–150}.
\newblock
\urldef\tempurl%
\url{https://doi.org/10.1145/15886.15902}
\showDOI{\tempurl}


\bibitem[Karl\'{\i}k et~al\mbox{.}(2019)]%
        {Karlik:2019:MISC}
\bibfield{author}{\bibinfo{person}{Ond\v{r}ej Karl\'{\i}k},
  \bibinfo{person}{Martin \v{S}ik}, \bibinfo{person}{Petr V\'{e}voda},
  \bibinfo{person}{Tom\'{a}\v{s} Sk\v{r}ivan}, {and} \bibinfo{person}{Jaroslav
  K\v{r}iv\'{a}nek}.} \bibinfo{year}{2019}\natexlab{}.
\newblock \showarticletitle{MIS Compensation: Optimizing Sampling Techniques in
  Multiple Importance Sampling}.
\newblock \bibinfo{journal}{\emph{ACM Transactions on Graphics}}
  \bibinfo{volume}{38}, \bibinfo{number}{6}, Article \bibinfo{articleno}{151}
  (\bibinfo{year}{2019}).
\newblock
\urldef\tempurl%
\url{https://doi.org/10.1145/3355089.3356565}
\showDOI{\tempurl}


\bibitem[Kingma and Dhariwal(2018)]%
        {Kingma:2018:Glow}
\bibfield{author}{\bibinfo{person}{Durk~P. Kingma} {and}
  \bibinfo{person}{Prafulla Dhariwal}.} \bibinfo{year}{2018}\natexlab{}.
\newblock \showarticletitle{Glow: Generative Flow with Invertible 1x1
  Convolutions}. In \bibinfo{booktitle}{\emph{Proceedings of the 32nd
  International Conference on Neural Information Processing Systems}}.
\newblock


\bibitem[Kuznetsov et~al\mbox{.}(2021)]%
        {Kuznetsov:2021:NeuMIP}
\bibfield{author}{\bibinfo{person}{Alexandr Kuznetsov},
  \bibinfo{person}{Krishna Mullia}, \bibinfo{person}{Zexiang Xu},
  \bibinfo{person}{Milo\v{s} Ha\v{s}an}, {and} \bibinfo{person}{Ravi
  Ramamoorthi}.} \bibinfo{year}{2021}\natexlab{}.
\newblock \showarticletitle{NeuMIP: Multi-resolution Neural Materials}.
\newblock \bibinfo{journal}{\emph{ACM Transactions on Graphics}}
  \bibinfo{volume}{40}, \bibinfo{number}{4} (\bibinfo{year}{2021}).
\newblock
\urldef\tempurl%
\url{https://doi.org/10.1145/3450626.3459795}
\showDOI{\tempurl}


\bibitem[Kuznetsov et~al\mbox{.}(2022)]%
        {Kuznetsov:2022:NeuMIP2}
\bibfield{author}{\bibinfo{person}{Alexandr Kuznetsov},
  \bibinfo{person}{Xuezheng Wang}, \bibinfo{person}{Krishna Mullia},
  \bibinfo{person}{Fujun Luan}, \bibinfo{person}{Zexiang Xu},
  \bibinfo{person}{Milo\v{s} Ha\v{s}an}, {and} \bibinfo{person}{Ravi
  Ramamoorthi}.} \bibinfo{year}{2022}\natexlab{}.
\newblock \showarticletitle{Rendering Neural materials on Curved Surfaces}. In
  \bibinfo{booktitle}{\emph{ACM SIGGRAPH 2022 Conference Proceedings}}.
  \bibinfo{pages}{1--9}.
\newblock
\urldef\tempurl%
\url{https://doi.org/10.1145/3528233.3530721}
\showDOI{\tempurl}


\bibitem[Loshchilov and Hutter(2019)]%
        {Loshchilov:2019:AdamW}
\bibfield{author}{\bibinfo{person}{Ilya Loshchilov} {and}
  \bibinfo{person}{Frank Hutter}.} \bibinfo{year}{2019}\natexlab{}.
\newblock \showarticletitle{Decoupled Weight Decay Regularization}. In
  \bibinfo{booktitle}{\emph{7th International Conference on Learning
  Representations}}.
\newblock


\bibitem[M\"uller(2021)]%
        {Muller:2021:tiny-cuda-nn}
\bibfield{author}{\bibinfo{person}{Thomas M\"uller}.}
  \bibinfo{year}{2021}\natexlab{}.
\newblock \bibinfo{booktitle}{\emph{{tiny-cuda-nn}}}.
\newblock
\urldef\tempurl%
\url{https://github.com/NVlabs/tiny-cuda-nn}
\showURL{%
\tempurl}


\bibitem[M{\"u}ller et~al\mbox{.}(2019)]%
        {Muller:2019:NIS}
\bibfield{author}{\bibinfo{person}{Thomas M{\"u}ller}, \bibinfo{person}{Brian
  McWilliams}, \bibinfo{person}{Fabrice Rousselle}, \bibinfo{person}{Markus
  Gross}, {and} \bibinfo{person}{Jan Nov{\'a}k}.}
  \bibinfo{year}{2019}\natexlab{}.
\newblock \showarticletitle{Neural Importance Sampling}.
\newblock \bibinfo{journal}{\emph{ACM Transactions on Graphics}}
  \bibinfo{volume}{38}, \bibinfo{number}{5} (\bibinfo{year}{2019}).
\newblock
\urldef\tempurl%
\url{https://doi.org/10.1145/3341156}
\showDOI{\tempurl}


\bibitem[M{\"u}ller et~al\mbox{.}(2020)]%
        {Muller:2020:NCV}
\bibfield{author}{\bibinfo{person}{Thomas M{\"u}ller}, \bibinfo{person}{Fabrice
  Rousselle}, \bibinfo{person}{Alexander Keller}, {and} \bibinfo{person}{Jan
  Nov{\'a}k}.} \bibinfo{year}{2020}\natexlab{}.
\newblock \showarticletitle{Neural Control Variates}.
\newblock \bibinfo{journal}{\emph{ACM Transactions on Graphics}}
  \bibinfo{volume}{39}, \bibinfo{number}{6} (\bibinfo{year}{2020}),
  \bibinfo{pages}{1--19}.
\newblock
\urldef\tempurl%
\url{https://doi.org/10.1145/3414685.3417804}
\showDOI{\tempurl}


\bibitem[Papamakarios et~al\mbox{.}(2021)]%
        {Papamakarios:2021:NF}
\bibfield{author}{\bibinfo{person}{George Papamakarios}, \bibinfo{person}{Eric
  Nalisnick}, \bibinfo{person}{Danilo~Jimenez Rezende}, \bibinfo{person}{Shakir
  Mohamed}, {and} \bibinfo{person}{Balaji Lakshminarayanan}.}
  \bibinfo{year}{2021}\natexlab{}.
\newblock \showarticletitle{Normalizing Flows for Probabilistic Modeling and
  Inference}.
\newblock \bibinfo{journal}{\emph{Journal of Machine Learning Research}}
  \bibinfo{volume}{22}, \bibinfo{number}{1}, Article \bibinfo{articleno}{57}
  (\bibinfo{year}{2021}).
\newblock
\urldef\tempurl%
\url{https://doi.org/10.5555/3546258.3546315}
\showDOI{\tempurl}


\bibitem[Paszke et~al\mbox{.}(2019)]%
        {Pytorch}
\bibfield{author}{\bibinfo{person}{Adam Paszke}, \bibinfo{person}{Sam Gross},
  \bibinfo{person}{Francisco Massa}, \bibinfo{person}{Adam Lerer},
  \bibinfo{person}{James Bradbury}, \bibinfo{person}{Gregory Chanan},
  \bibinfo{person}{Trevor Killeen}, \bibinfo{person}{Zeming Lin},
  \bibinfo{person}{Natalia Gimelshein}, \bibinfo{person}{Luca Antiga},
  \bibinfo{person}{Alban Desmaison}, \bibinfo{person}{Andreas Kopf},
  \bibinfo{person}{Edward Yang}, \bibinfo{person}{Zachary DeVito},
  \bibinfo{person}{Martin Raison}, \bibinfo{person}{Alykhan Tejani},
  \bibinfo{person}{Sasank Chilamkurthy}, \bibinfo{person}{Benoit Steiner},
  \bibinfo{person}{Lu Fang}, \bibinfo{person}{Junjie Bai}, {and}
  \bibinfo{person}{Soumith Chintala}.} \bibinfo{year}{2019}\natexlab{}.
\newblock \bibinfo{booktitle}{\emph{PyTorch: An Imperative Style,
  High-Performance Deep Learning Library}}.
\newblock \bibinfo{pages}{8024--8035}.
\newblock
\urldef\tempurl%
\url{https://doi.org/10.5555/3454287.3455008}
\showDOI{\tempurl}


\bibitem[Rezende and Mohamed(2015)]%
        {Rezende:2015:NF}
\bibfield{author}{\bibinfo{person}{Danilo~Jimenez Rezende} {and}
  \bibinfo{person}{Shakir Mohamed}.} \bibinfo{year}{2015}\natexlab{}.
\newblock \showarticletitle{Variational Inference with Normalizing Flows}. In
  \bibinfo{booktitle}{\emph{Proceedings of the 32nd International Conference on
  International Conference on Machine Learning}}, Vol.~\bibinfo{volume}{37}.
  \bibinfo{pages}{1530--1538}.
\newblock


\bibitem[Rezende et~al\mbox{.}(2020)]%
        {Jimenez:2020:CircularFlows}
\bibfield{author}{\bibinfo{person}{Danilo~Jimenez Rezende},
  \bibinfo{person}{George Papamakarios}, \bibinfo{person}{Sebastien Racaniere},
  \bibinfo{person}{Michael Albergo}, \bibinfo{person}{Gurtej Kanwar},
  \bibinfo{person}{Phiala Shanahan}, {and} \bibinfo{person}{Kyle Cranmer}.}
  \bibinfo{year}{2020}\natexlab{}.
\newblock \showarticletitle{Normalizing Flows on Tori and Spheres}. In
  \bibinfo{booktitle}{\emph{Proceedings of the 37th International Conference on
  Machine Learning}}. \bibinfo{pages}{8083--8092}.
\newblock
\urldef\tempurl%
\url{https://doi.org/10.5555/3045118.3045281}
\showDOI{\tempurl}


\bibitem[Sztrajman et~al\mbox{.}(2021)]%
        {Ssztrajman:2021:NeuralBRDF}
\bibfield{author}{\bibinfo{person}{Alejandro Sztrajman},
  \bibinfo{person}{Gilles Rainer}, \bibinfo{person}{Tobias Ritschel}, {and}
  \bibinfo{person}{Tim Weyrich}.} \bibinfo{year}{2021}\natexlab{}.
\newblock \showarticletitle{Neural BRDF Representation and Importance
  Sampling}. In \bibinfo{booktitle}{\emph{Computer Graphics Forum}},
  Vol.~\bibinfo{volume}{40}. \bibinfo{pages}{332--346}.
\newblock
\urldef\tempurl%
\url{https://doi.org/10.1111/cgf.14335}
\showDOI{\tempurl}


\bibitem[Talbot(2005)]%
        {Talbot:2005:Thesis}
\bibfield{author}{\bibinfo{person}{Justin~F. Talbot}.}
  \bibinfo{year}{2005}\natexlab{}.
\newblock \bibinfo{title}{Importance Resampling for Global Illumination}.
\newblock
\newblock


\bibitem[Talbot et~al\mbox{.}(2005)]%
        {Talbot:2005:RIS}
\bibfield{author}{\bibinfo{person}{Justin~F. Talbot}, \bibinfo{person}{David
  Cline}, {and} \bibinfo{person}{Parris Egbert}.}
  \bibinfo{year}{2005}\natexlab{}.
\newblock \showarticletitle{Importance Resampling for Global Illumination}. In
  \bibinfo{booktitle}{\emph{Proceedings of the 16th Eurographics Conference on
  Rendering Techniques}} \emph{(\bibinfo{series}{EGSR '05})}.
  \bibinfo{pages}{139–146}.
\newblock
\urldef\tempurl%
\url{https://doi.org/10.5555/2383654.2383674}
\showDOI{\tempurl}


\bibitem[Tong et~al\mbox{.}(2024)]%
        {Tong:2024:Improving}
\bibfield{author}{\bibinfo{person}{Alexander Tong}, \bibinfo{person}{Kilian
  Fatras}, \bibinfo{person}{Nikolay Malkin}, \bibinfo{person}{Guillaume
  Huguet}, \bibinfo{person}{Yanlei Zhang}, \bibinfo{person}{Jarrid
  Rector-Brooks}, \bibinfo{person}{Guy Wolf}, {and} \bibinfo{person}{Yoshua
  Bengio}.} \bibinfo{year}{2024}\natexlab{}.
\newblock \showarticletitle{Improving and Generalizing Flow-based Generative
  Models with Minibatch Optimal Transport}.
\newblock \bibinfo{journal}{\emph{Transactions on Machine Learning Research}}
  (\bibinfo{year}{2024}).
\newblock


\bibitem[Vaidyanathan et~al\mbox{.}(2023)]%
        {Vaidyanathan:2023:Random}
\bibfield{author}{\bibinfo{person}{Karthik Vaidyanathan},
  \bibinfo{person}{Marco Salvi}, \bibinfo{person}{Bartlomiej Wronski},
  \bibinfo{person}{Tomas Akenine-M\"oller}, \bibinfo{person}{Pontus Ebelin},
  {and} \bibinfo{person}{Aaron Lefohn}.} \bibinfo{year}{2023}\natexlab{}.
\newblock \showarticletitle{Random-Access Neural Compression of Material
  Textures}.
\newblock \bibinfo{journal}{\emph{ACM Transactions on Graphics}}
  \bibinfo{volume}{42}, \bibinfo{number}{4}, Article \bibinfo{articleno}{88}
  (\bibinfo{date}{July} \bibinfo{year}{2023}).
\newblock
\urldef\tempurl%
\url{https://doi.org/10.1145/3592407}
\showDOI{\tempurl}


\bibitem[Veach and Guibas(1995)]%
        {Veach:1995:MIS}
\bibfield{author}{\bibinfo{person}{Eric Veach} {and}
  \bibinfo{person}{Leonidas~J. Guibas}.} \bibinfo{year}{1995}\natexlab{}.
\newblock \showarticletitle{Optimally Combining Sampling Techniques for Monte
  Carlo Rendering}. In \bibinfo{booktitle}{\emph{Proceedings of the 22nd Annual
  Conference on Computer Graphics and Interactive Techniques}}
  \emph{(\bibinfo{series}{SIGGRAPH '95})}. \bibinfo{pages}{419–428}.
\newblock
\urldef\tempurl%
\url{https://doi.org/10.1145/218380.218498}
\showDOI{\tempurl}


\bibitem[Vicini et~al\mbox{.}(2019)]%
        {Vicini:2019:Learned}
\bibfield{author}{\bibinfo{person}{Delio Vicini}, \bibinfo{person}{Vladlen
  Koltun}, {and} \bibinfo{person}{Wenzel Jakob}.}
  \bibinfo{year}{2019}\natexlab{}.
\newblock \showarticletitle{A Learned Shape-Adaptive Subsurface Scattering
  Model}.
\newblock \bibinfo{journal}{\emph{ACM Transactions on Graphics}}
  \bibinfo{volume}{38}, \bibinfo{number}{4} (\bibinfo{date}{July}
  \bibinfo{year}{2019}), \bibinfo{pages}{15}.
\newblock
\urldef\tempurl%
\url{https://doi.org/10.1145/3306346.3322974}
\showDOI{\tempurl}


\bibitem[Villeneuve et~al\mbox{.}(2021)]%
        {Villeneuve:2021:VolumeProductSampling}
\bibfield{author}{\bibinfo{person}{Keven Villeneuve}, \bibinfo{person}{Adrien
  Gruson}, \bibinfo{person}{Iliyan Georgiev}, {and} \bibinfo{person}{Derek
  Nowrouzezahrai}.} \bibinfo{year}{2021}\natexlab{}.
\newblock \showarticletitle{Practical Product Sampling for Single Scattering in
  Media}. In \bibinfo{booktitle}{\emph{Proceedings of EGSR}}.
\newblock
\urldef\tempurl%
\url{https://doi.org/10.2312/sr.20211290}
\showDOI{\tempurl}


\bibitem[Xia et~al\mbox{.}(2020)]%
        {Xia:2019:Product}
\bibfield{author}{\bibinfo{person}{Mengqi~(Mandy) Xia}, \bibinfo{person}{Bruce
  Walter}, \bibinfo{person}{Christophe Hery}, {and} \bibinfo{person}{Steve
  Marschner}.} \bibinfo{year}{2020}\natexlab{}.
\newblock \showarticletitle{Gaussian Product Sampling for Rendering Layered
  Materials}.
\newblock \bibinfo{journal}{\emph{Computer Graphics Forum}}
  \bibinfo{volume}{39}, \bibinfo{number}{1} (\bibinfo{year}{2020}),
  \bibinfo{pages}{420--435}.
\newblock
\urldef\tempurl%
\url{https://doi.org/10.1111/cgf.13883}
\showDOI{\tempurl}


\bibitem[Xu et~al\mbox{.}(2023)]%
        {Xu:2023:NeuSample}
\bibfield{author}{\bibinfo{person}{Bing Xu}, \bibinfo{person}{Liwen Wu},
  \bibinfo{person}{Miloš Hašan}, \bibinfo{person}{Fujun Luan},
  \bibinfo{person}{Iliyan Georgiev}, \bibinfo{person}{Zexiang Xu}, {and}
  \bibinfo{person}{Ravi Ramamoorthi}.} \bibinfo{year}{2023}\natexlab{}.
\newblock \showarticletitle{NeuSample: Importance Sampling for Neural
  Materials}. In \bibinfo{booktitle}{\emph{ACM SIGGRAPH 2023 Conference
  Proceedings}}.
\newblock
\showISBNx{979-8-4007-0159-7/23/08}
\urldef\tempurl%
\url{https://doi.org/10.1145/3588432.3591524}
\showDOI{\tempurl}


\bibitem[Zeltner et~al\mbox{.}(2024)]%
        {Zeltner:2023:Real}
\bibfield{author}{\bibinfo{person}{Tizian Zeltner}, \bibinfo{person}{Fabrice
  Rousselle}, \bibinfo{person}{Andrea Weidlich}, \bibinfo{person}{Petrik
  Clarberg}, \bibinfo{person}{Jan Nov{\'a}k}, \bibinfo{person}{Benedikt
  Bitterli}, \bibinfo{person}{Alex Evans}, \bibinfo{person}{Tom{\'a}{\v{s}}
  Davidovi{\v{c}}}, \bibinfo{person}{Simon Kallweit}, {and}
  \bibinfo{person}{Aaron Lefohn}.} \bibinfo{year}{2024}\natexlab{}.
\newblock \showarticletitle{Real-Time Neural Appearance Models}.
\newblock \bibinfo{journal}{\emph{ACM Transactions on Graphics}}
  (\bibinfo{year}{2024}).
\newblock
\urldef\tempurl%
\url{https://doi.org/10.1145/3659577}
\showDOI{\tempurl}


\bibitem[Zheng and Zwicker(2019)]%
        {Zheng:2019:Learning}
\bibfield{author}{\bibinfo{person}{Quan Zheng} {and} \bibinfo{person}{Matthias
  Zwicker}.} \bibinfo{year}{2019}\natexlab{}.
\newblock \showarticletitle{Learning to Importance Sample in Primary Sample
  Space}. In \bibinfo{booktitle}{\emph{Computer Graphics Forum}},
  Vol.~\bibinfo{volume}{38}. \bibinfo{pages}{169--179}.
\newblock
\urldef\tempurl%
\url{https://doi.org/10.1111/cgf.13628}
\showDOI{\tempurl}


\end{thebibliography}

\newpage

\end{document}